\documentclass[letterpaper, 10 pt, conference]{ieeeconf}  

\IEEEoverridecommandlockouts                              

\usepackage{macros}
\usepackage{gensymb}

\usepackage{xspace}
\usepackage{bm}

\usepackage{cite}
\usepackage{amsmath,amssymb,amsfonts}
\usepackage{graphicx}
\usepackage{textcomp}
\usepackage{xcolor}
\usepackage{comment}
\def\BibTeX{{\rm B\kern-.05em{\sc i\kern-.025em b}\kern-.08em
    T\kern-.1667em\lower.7ex\hbox{E}\kern-.125emX}}
\usepackage{hyperref}
\usepackage{subcaption}
\usepackage{ragged2e}
\usepackage[noend]{algpseudocode}
\usepackage{algorithm}

\newcommand{\safemath}[2]{\newcommand{#1}{\ensuremath{#2}\xspace}}

\safemath{\bma}{\mathbf{a}}
\safemath{\bmb}{\mathbf{b}}
\safemath{\bmc}{\mathbf{c}}
\safemath{\bmd}{\mathbf{d}}
\safemath{\bme}{\mathbf{e}}
\safemath{\bmf}{\mathbf{f}}
\safemath{\bmg}{\mathbf{g}}
\safemath{\bmh}{\mathbf{h}}
\safemath{\bmi}{\mathbf{i}}
\safemath{\bmj}{\mathbf{j}}
\safemath{\bmk}{\mathbf{k}}
\safemath{\bml}{\mathbf{l}}
\safemath{\bmm}{\mathbf{m}}
\safemath{\bmn}{\mathbf{n}}
\safemath{\bmo}{\mathbf{o}}
\safemath{\bmp}{\mathbf{p}}
\safemath{\bmq}{\mathbf{q}}
\safemath{\bmr}{\mathbf{r}}
\safemath{\bms}{\mathbf{s}}
\safemath{\bmt}{\mathbf{t}}
\safemath{\bmu}{\mathbf{u}}
\safemath{\bmv}{\mathbf{v}}
\safemath{\bmw}{\mathbf{w}}
\safemath{\bmx}{\mathbf{x}}
\safemath{\bmy}{\mathbf{y}}
\safemath{\bmz}{\mathbf{z}}
\safemath{\bmzero}{\mathbf{0}}
\safemath{\bmone}{\mathbf{1}}

\bmdefine{\biad}{a}
\bmdefine{\bibd}{b}
\bmdefine{\bicd}{c}
\bmdefine{\bidd}{d}
\bmdefine{\bied}{e}
\bmdefine{\bifd}{f}
\bmdefine{\bigd}{g}
\bmdefine{\bihd}{h}
\bmdefine{\biid}{i}
\bmdefine{\bijd}{j}
\bmdefine{\bikd}{k}
\bmdefine{\bild}{l}
\bmdefine{\bimd}{m}
\bmdefine{\bind}{n}
\bmdefine{\biod}{o}
\bmdefine{\bipd}{p}
\bmdefine{\biqd}{q}
\bmdefine{\bird}{r}
\bmdefine{\bisd}{s}
\bmdefine{\bitd}{t}
\bmdefine{\biud}{u}
\bmdefine{\bivd}{v}
\bmdefine{\biwd}{w}
\bmdefine{\bixd}{x}
\bmdefine{\biyd}{y}
\bmdefine{\bizd}{z}

\bmdefine{\bixid}{\xi}
\bmdefine{\bilambdad}{\lambda}
\bmdefine{\bimud}{\mu}
\bmdefine{\bithetad}{\theta}
\bmdefine{\biphid}{\phi}
\bmdefine{\bideltad}{\delta}

\safemath{\bmia}{\biad}
\safemath{\bmib}{\bibd}
\safemath{\bmic}{\bicd}
\safemath{\bmid}{\bidd}
\safemath{\bmie}{\bied}
\safemath{\bmif}{\bifd}
\safemath{\bmig}{\bigd}
\safemath{\bmih}{\bihd}
\safemath{\bmii}{\biid}
\safemath{\bmij}{\bijd}
\safemath{\bmik}{\bikd}
\safemath{\bmil}{\bild}
\safemath{\bmim}{\bimd}
\safemath{\bmin}{\bind}
\safemath{\bmio}{\biod}
\safemath{\bmip}{\bipd}
\safemath{\bmiq}{\biqd}
\safemath{\bmir}{\bird}
\safemath{\bmis}{\bisd}
\safemath{\bmit}{\bitd}
\safemath{\bmiu}{\biud}
\safemath{\bmiv}{\bivd}
\safemath{\bmiw}{\biwd}
\safemath{\bmix}{\bixd}
\safemath{\bmiy}{\biyd}
\safemath{\bmiz}{\bizd}

\safemath{\bmxi}{\bixid}
\safemath{\bmlambda}{\bilambdad}
\safemath{\bmmu}{\bimud}
\safemath{\bmtheta}{\bithetad}
\safemath{\bmphi}{\biphid}
\safemath{\bmdelta}{\bideltad}

\safemath{\bA}{\mathbf{A}}
\safemath{\bB}{\mathbf{B}}
\safemath{\bC}{\mathbf{C}}
\safemath{\bD}{\mathbf{D}}
\safemath{\bE}{\mathbf{E}}
\safemath{\bF}{\mathbf{F}}
\safemath{\bG}{\mathbf{G}}
\safemath{\bH}{\mathbf{H}}
\safemath{\bI}{\mathbf{I}}
\safemath{\bJ}{\mathbf{J}}
\safemath{\bK}{\mathbf{K}}
\safemath{\bL}{\mathbf{L}}
\safemath{\bM}{\mathbf{M}}
\safemath{\bN}{\mathbf{N}}
\safemath{\bO}{\mathbf{O}}
\safemath{\bP}{\mathbf{P}}
\safemath{\bQ}{\mathbf{Q}}
\safemath{\bR}{\mathbf{R}}
\safemath{\bS}{\mathbf{S}}
\safemath{\bT}{\mathbf{T}}
\safemath{\bU}{\mathbf{U}}
\safemath{\bV}{\mathbf{V}}
\safemath{\bW}{\mathbf{W}}
\safemath{\bX}{\mathbf{X}}
\safemath{\bY}{\mathbf{Y}}
\safemath{\bZ}{\mathbf{Z}}

\safemath{\bZero}{\mathbf{0}}
\safemath{\bOne}{\mathbf{1}}
\safemath{\bDelta}{\mathbf{\Delta}}
\safemath{\bLambda}{\boldsymbol\Lambda}
\safemath{\bPhi}{\mathbf{\Upphi}}
\safemath{\bSigma}{\mathbf{\Upsigma}}
\safemath{\bOmega}{\mathbf{\Upomega}}
\safemath{\bTheta}{\mathbf{\Uptheta}}

\bmdefine{\biAd}{A}
\bmdefine{\biBd}{B}
\bmdefine{\biCd}{C}
\bmdefine{\biDd}{D}
\bmdefine{\biEd}{E}
\bmdefine{\biFd}{F}
\bmdefine{\biGd}{G}
\bmdefine{\biHd}{H}
\bmdefine{\biId}{I}
\bmdefine{\biJd}{J}
\bmdefine{\biKd}{K}
\bmdefine{\biLd}{L}
\bmdefine{\biMd}{M}
\bmdefine{\biOd}{N}
\bmdefine{\biPd}{O}
\bmdefine{\biQd}{P}
\bmdefine{\biRd}{R}
\bmdefine{\biSd}{S}
\bmdefine{\biTd}{T}
\bmdefine{\biUd}{U}
\bmdefine{\biVd}{V}
\bmdefine{\biWd}{W}
\bmdefine{\biXd}{X}
\bmdefine{\biYd}{Y}
\bmdefine{\biZd}{Z}

\bmdefine{\biDelta}{\Delta}
\bmdefine{\biLambda}{\Lambda}
\bmdefine{\biPhi}{\Phi}
\bmdefine{\biSigma}{\Sigma}
\bmdefine{\biOmega}{\Omega}
\bmdefine{\biTheta}{\Theta}

\safemath{\bimA}{\biAd}
\safemath{\bimB}{\biBd}
\safemath{\bimC}{\biCd}
\safemath{\bimD}{\biDd}
\safemath{\bimE}{\biEd}
\safemath{\bimF}{\biFd}
\safemath{\bimG}{\biGd}
\safemath{\bimH}{\biHd}
\safemath{\bimI}{\biId}
\safemath{\bimJ}{\biJd}
\safemath{\bimK}{\biKd}
\safemath{\bimL}{\biLd}
\safemath{\bimM}{\biMd}
\safemath{\bimN}{\biNd}
\safemath{\bimO}{\biOd}
\safemath{\bimP}{\biPd}
\safemath{\bimQ}{\biQd}
\safemath{\bimR}{\biRd}
\safemath{\bimS}{\biSd}
\safemath{\bimT}{\biTd}
\safemath{\bimU}{\biUd}
\safemath{\bimV}{\biVd}
\safemath{\bimW}{\biWd}
\safemath{\bimX}{\biXd}
\safemath{\bimY}{\biYd}
\safemath{\bimZ}{\biZd}

\safemath{\bimDelta}{\biDelta}
\safemath{\bimLambda}{\biLambda}
\safemath{\bimPhi}{\biPhi}
\safemath{\bimSigma}{\biSigma}
\safemath{\bimOmega}{\biOmega}
\safemath{\bimTheta}{\biTheta}

\safemath{\setA}{\mathcal{A}}
\safemath{\setB}{\mathcal{B}}
\safemath{\setC}{\mathcal{C}}
\safemath{\setD}{\mathcal{D}}
\safemath{\setE}{\mathcal{E}}
\safemath{\setF}{\mathcal{F}}
\safemath{\setG}{\mathcal{G}}
\safemath{\setH}{\mathcal{H}}
\safemath{\setI}{\mathcal{I}}
\safemath{\setJ}{\mathcal{J}}
\safemath{\setK}{\mathcal{K}}
\safemath{\setL}{\mathcal{L}}
\safemath{\setM}{\mathcal{M}}
\safemath{\setN}{\mathcal{N}}
\safemath{\setO}{\mathcal{O}}
\safemath{\setP}{\mathcal{P}}
\safemath{\setQ}{\mathcal{Q}}
\safemath{\setR}{\mathcal{R}}
\safemath{\setS}{\mathcal{S}}
\safemath{\setT}{\mathcal{T}}
\safemath{\setU}{\mathcal{U}}
\safemath{\setW}{\mathcal{W}}
\safemath{\setX}{\mathcal{X}}
\safemath{\setY}{\mathcal{Y}}
\safemath{\setZ}{\mathcal{Z}}
\safemath{\emptySet}{\varnothing}

\safemath{\colA}{\mathscr{A}}
\safemath{\colB}{\mathscr{B}}
\safemath{\colC}{\mathscr{C}}
\safemath{\colD}{\mathscr{D}}
\safemath{\colE}{\mathscr{E}}
\safemath{\colF}{\mathscr{F}}
\safemath{\colG}{\mathscr{G}}
\safemath{\colH}{\mathscr{H}}
\safemath{\colI}{\mathscr{I}}
\safemath{\colJ}{\mathscr{J}}
\safemath{\colK}{\mathscr{K}}
\safemath{\colL}{\mathscr{L}}
\safemath{\colM}{\mathscr{M}}
\safemath{\colN}{\mathscr{N}}
\safemath{\colO}{\mathscr{O}}
\safemath{\colP}{\mathscr{P}}
\safemath{\colQ}{\mathscr{Q}}
\safemath{\colR}{\mathscr{R}}
\safemath{\colS}{\mathscr{S}}
\safemath{\colT}{\mathscr{T}}
\safemath{\colU}{\mathscr{U}}
\safemath{\colV}{\mathscr{V}}
\safemath{\colW}{\mathscr{W}}
\safemath{\colX}{\mathscr{X}}
\safemath{\colY}{\mathscr{Y}}
\safemath{\colZ}{\mathscr{Z}}

\safemath{\opA}{\mathbb{A}}
\safemath{\opB}{\mathbb{B}}
\safemath{\opC}{\mathbb{C}}
\safemath{\opD}{\mathbb{D}}
\safemath{\opE}{\mathbb{E}}
\safemath{\opF}{\mathbb{F}}
\safemath{\opG}{\mathbb{G}}
\safemath{\opH}{\mathbb{H}}
\safemath{\opI}{\mathbb{I}}
\safemath{\opJ}{\mathbb{J}}
\safemath{\opK}{\mathbb{K}}
\safemath{\opL}{\mathbb{L}}
\safemath{\opM}{\mathbb{M}}
\safemath{\opN}{\mathbb{N}}
\safemath{\opO}{\mathbb{O}}
\safemath{\opP}{\mathbb{P}}
\safemath{\opQ}{\mathbb{Q}}
\safemath{\opR}{\mathbb{R}}
\safemath{\opS}{\mathbb{S}}
\safemath{\opT}{\mathbb{T}}
\safemath{\opU}{\mathbb{U}}
\safemath{\opV}{\mathbb{V}}
\safemath{\opW}{\mathbb{W}}
\safemath{\opX}{\mathbb{X}}
\safemath{\opY}{\mathbb{Y}}
\safemath{\opZ}{\mathbb{Z}}
\safemath{\opZero}{\mathbb{O}}
\safemath{\identityop}{\opI}

\safemath{\veca}{\bma}
\safemath{\vecb}{\bmb}
\safemath{\vecc}{\bmc}
\safemath{\vecd}{\bmd}
\safemath{\vece}{\bme}
\safemath{\vecf}{\bmf}
\safemath{\vecg}{\bmg}
\safemath{\vech}{\bmh}
\safemath{\veci}{\bmi}
\safemath{\vecj}{\bmj}
\safemath{\veck}{\bmk}
\safemath{\vecl}{\bml}
\safemath{\vecm}{\bmm}
\safemath{\vecn}{\bmn}
\safemath{\veco}{\bmo}
\safemath{\vecp}{\bmp}
\safemath{\vecq}{\bmq}
\safemath{\vecr}{\bmr}
\safemath{\vecs}{\bms}
\safemath{\vect}{\bmt}
\safemath{\vecu}{\bmu}
\safemath{\vecv}{\bmv}
\safemath{\vecw}{\bmw}
\safemath{\vecx}{\bmx}
\safemath{\vecy}{\bmy}
\safemath{\vecz}{\bmz}

\safemath{\veczero}{\bmzero}
\safemath{\vecone}{\bmone}
\safemath{\vecxi}{\bmxi}
\safemath{\veclambda}{\bmlambda}
\safemath{\vecmu}{\bmmu}
\safemath{\vectheta}{\bmtheta}
\safemath{\vecphi}{\bmphi}
\safemath{\vecdelta}{\bmdelta}

\safemath{\matA}{\bA}
\safemath{\matB}{\bB}
\safemath{\matC}{\bC}
\safemath{\matD}{\bD}
\safemath{\matE}{\bE}
\safemath{\matF}{\bF}
\safemath{\matG}{\bG}
\safemath{\matH}{\bH}
\safemath{\matI}{\bI}
\safemath{\matJ}{\bJ}
\safemath{\matK}{\bK}
\safemath{\matL}{\bL}
\safemath{\matM}{\bM}
\safemath{\matN}{\bN}
\safemath{\matO}{\bO}
\safemath{\matP}{\bP}
\safemath{\matQ}{\bQ}
\safemath{\matR}{\bR}
\safemath{\matS}{\bS}
\safemath{\matT}{\bT}
\safemath{\matU}{\bU}
\safemath{\matV}{\bV}
\safemath{\matW}{\bW}
\safemath{\matX}{\bX}
\safemath{\matY}{\bY}
\safemath{\matZ}{\bZ}
\safemath{\matzero}{\bmzero}

\safemath{\matDelta}{\bDelta}
\safemath{\matLambda}{\bLambda}
\safemath{\matPhi}{\bPhi}
\safemath{\matSigma}{\bSigma}
\safemath{\matOmega}{\bOmega}
\safemath{\matTheta}{\bTheta}

\safemath{\matidentity}{\matI}
\safemath{\matone}{\matO}

\safemath{\rnda}{A}
\safemath{\rndb}{B}
\safemath{\rndc}{C}
\safemath{\rndd}{D}
\safemath{\rnde}{E}
\safemath{\rndf}{F}
\safemath{\rndg}{G}
\safemath{\rndh}{H}
\safemath{\rndi}{I}
\safemath{\rndj}{J}
\safemath{\rndk}{K}
\safemath{\rndl}{L}
\safemath{\rndm}{M}
\safemath{\rndn}{N}
\safemath{\rndo}{O}
\safemath{\rndp}{P}
\safemath{\rndq}{Q}
\safemath{\rndr}{R}
\safemath{\rnds}{S}
\safemath{\rndt}{T}
\safemath{\rndu}{U}
\safemath{\rndv}{V}
\safemath{\rndw}{W}
\safemath{\rndx}{X}
\safemath{\rndy}{Y}
\safemath{\rndz}{Z}

\safemath{\rveca}{\bimA}
\safemath{\rvecb}{\bimB}
\safemath{\rvecc}{\bimC}
\safemath{\rvecd}{\bimD}
\safemath{\rvece}{\bimE}
\safemath{\rvecf}{\bimF}
\safemath{\rvecg}{\bimG}
\safemath{\rvech}{\bimH}
\safemath{\rveci}{\bimI}
\safemath{\rvecj}{\bimJ}
\safemath{\rveck}{\bimK}
\safemath{\rvecl}{\bimL}
\safemath{\rvecm}{\bimM}
\safemath{\rvecn}{\bimN}
\safemath{\rveco}{\bomO}
\safemath{\rvecp}{\bimP}
\safemath{\rvecq}{\bimQ}
\safemath{\rvecr}{\bimR}
\safemath{\rvecs}{\bimS}
\safemath{\rvect}{\bimT}
\safemath{\rvecu}{\bimU}
\safemath{\rvecv}{\bimV}
\safemath{\rvecw}{\bimW}
\safemath{\rvecx}{\bimX}
\safemath{\rvecy}{\bimY}
\safemath{\rvecz}{\bimZ}

\safemath{\rvecxi}{\bmxi}
\safemath{\rveclambda}{\bmlambda}
\safemath{\rvecmu}{\bmmu}
\safemath{\rvectheta}{\bmtheta}
\safemath{\rvecphi}{\bmphi}

\safemath{\rmatA}{\bimA}
\safemath{\rmatB}{\bimB}
\safemath{\rmatC}{\bimC}
\safemath{\rmatD}{\bimD}
\safemath{\rmatE}{\bimE}
\safemath{\rmatF}{\bimF}
\safemath{\rmatG}{\bimG}
\safemath{\rmatH}{\bimH}
\safemath{\rmatI}{\bimI}
\safemath{\rmatJ}{\bimJ}
\safemath{\rmatK}{\bimK}
\safemath{\rmatL}{\bimL}
\safemath{\rmatM}{\bimM}
\safemath{\rmatN}{\bimN}
\safemath{\rmatO}{\bimO}
\safemath{\rmatP}{\bimP}
\safemath{\rmatQ}{\bimQ}
\safemath{\rmatR}{\bimR}
\safemath{\rmatS}{\bimS}
\safemath{\rmatT}{\bimT}
\safemath{\rmatU}{\bimU}
\safemath{\rmatV}{\bimV}
\safemath{\rmatW}{\bimW}
\safemath{\rmatX}{\bimX}
\safemath{\rmatY}{\bimY}
\safemath{\rmatZ}{\bimZ}

\safemath{\rmatDelta}{\bimDelta}
\safemath{\rmatLambda}{\bimLambda}
\safemath{\rmatPhi}{\bimPhi}
\safemath{\rmatSigma}{\bimSigma}
\safemath{\rmatOmega}{\bimOmega}
\safemath{\rmatTheta}{\bimTheta}

\usepackage{amssymb}
\usepackage{amsfonts}
\usepackage{mathrsfs}
\usepackage{xspace}
\usepackage{bm}
\usepackage{fancyref}
\usepackage{textcomp}

\usepackage{multirow}
\usepackage{stmaryrd}

\newenvironment{textbmatrix}{	\setlength{\arraycolsep}{2.5pt}%
								\big[\begin{matrix}}{\end{matrix}\big]%
								\raisebox{0.08ex}{\vphantom{M}}}

\def\be{\begin{equation}}
\def\ee{\end{equation}}
\def\een{\nonumber \end{equation}}
\def\mat{\begin{bmatrix}}
\def\emat{\end{bmatrix}}
\def\btm{\begin{textbmatrix}}
\def\etm{\end{textbmatrix}}

\def\ba#1\ea{\begin{align}#1\end{align}}
\def\bas#1\eas{\begin{align*}#1\end{align*}}
\def\bs#1\es{\begin{split}#1\end{split}} 
\def\bg#1\eg{\begin{gather}#1\end{gather}}
\def\bml#1\eml{\begin{multline}#1\end{multline}}
\def\bi#1\ei{\begin{itemize}#1\end{itemize}}

\safemath{\dirac}{\delta}					
\safemath{\krond}{\dirac}					

\safemath{\upto}{\uparrow}
\safemath{\downto}{\downarrow}
\safemath{\iu}{j}							
\safemath{\hilseqspace}{l^{2}}				
\newcommand{\banachfunspace}[1]{\setL^{#1}}	
\safemath{\hilfunspace}{\banachfunspace{2}}	

\safemath{\SNR}{\textsf{SNR}} 				
\safemath{\PAR}{\textsf{PAR}} 				
\safemath{\No}{N_0}							
\safemath{\Es}{E_s}							
\safemath{\EbNo}{\frac{\Eb}{\No}}
\safemath{\EsNo}{\frac{\Es}{\No}}

\DeclareMathOperator{\CHop}{\ensuremath{\opH}} 
\safemath{\tvir}{\rndh_{\CHop}}				
\safemath{\tvtf}{\rndl_{\CHop}}				
\safemath{\spf}{\rnds_{\CHop}}				
\safemath{\bff}{H_{\CHop}}					

\safemath{\ircf}{r_{h}}						
\safemath{\tftvcf}{r_{s}}					
\safemath{\tfcf}{r_{l}}						
\safemath{\bfcf}{r_{H}}						

\safemath{\tcorr}{c_h}						
\safemath{\scf}{c_{s}}						
\safemath{\tfcorr}{c_{l}}					
\safemath{\fcorr}{c_{H}}						

\safemath{\mi}{I}							
\safemath{\capacity}{C}						

\safemath{\jpg}{\mathcal{CN}}			
\safemath{\mchain}{\leftrightarrow}		

\safemath{\dB}{\,\mathrm{dB}}
\safemath{\dBm}{\,\mathrm{dBm}}
\safemath{\Hz}{\,\mathrm{Hz}}
\safemath{\kHz}{\,\mathrm{kHz}}
\safemath{\MHz}{\,\mathrm{MHz}}
\safemath{\GHz}{\,\mathrm{GHz}}
\safemath{\s}{\,\mathrm{s}}
\safemath{\ms}{\,\mathrm{ms}}
\safemath{\mus}{\,\mathrm{\text{\textmu}s}}
\safemath{\ns}{\,\mathrm{ns}}
\safemath{\ps}{\,\mathrm{ps}}
\safemath{\meter}{\,\mathrm{m}}
\safemath{\mm}{\,\mathrm{mm}}
\safemath{\cm}{\,\mathrm{cm}}
\safemath{\m}{\,\mathrm{m}}
\safemath{\W}{\,\mathrm{W}}
\safemath{\mW}{\, \mathrm{mW}}
\safemath{\J}{\,\mathrm{J}}
\safemath{\K}{\,\mathrm{K}}
\safemath{\bit}{\,\mathrm{bit}}
\safemath{\nat}{\,\mathrm{nat}}

\safemath{\define}{\triangleq}			

\safemath{\equivalent}{\sim}
\safemath{\distas}{\sim}					
\safemath{\sdiff}{\Delta}				

\safemath{\reals}{\mathbb{R}}
\safemath{\positivereals}{\reals_{+}}
\safemath{\integers}{\mathbb{Z}}
\safemath{\posint}{\integers_{+}}
\safemath{\naturals}{\mathbb{N}}
\safemath{\posnaturals}{\naturals_{+}}
\safemath{\complexset}{\mathbb{C}}
\safemath{\rationals}{\mathbb{Q}}

\newcommand*{\fancyrefapplabelprefix}{app}		
\newcommand*{\fancyrefthmlabelprefix}{thm}		
\newcommand*{\fancyreflemlabelprefix}{lem}		
\newcommand*{\fancyrefcorlabelprefix}{cor}		
\newcommand*{\fancyrefdeflabelprefix}{def}		
\newcommand*{\fancyrefproplabelprefix}{prop}	
\newcommand*{\fancyrefobslabelprefix}{obs}		
\newcommand*{\fancyrefalglabelprefix}{alg}		
\newcommand*{\fancyrefasmlabelprefix}{asm}	    
\newcommand*{\fancyrefasmslabelprefix}{asms}	    
\newcommand*{\fancyreftbllabelprefix}{tbl}	    
\newcommand*{\fancyrefestilabelprefix}{esti}	    

\frefformat{vario}{\fancyrefseclabelprefix}{Section~#1}
\frefformat{vario}{\fancyrefthmlabelprefix}{Theorem~#1}
\frefformat{vario}{\fancyreflemlabelprefix}{Lemma~#1}
\frefformat{vario}{\fancyrefcorlabelprefix}{Corollary~#1}
\frefformat{vario}{\fancyrefdeflabelprefix}{Definition~#1}
\frefformat{vario}{\fancyrefobslabelprefix}{Observation~#1}
\frefformat{vario}{\fancyrefasmlabelprefix}{Assumption~#1}
\frefformat{vario}{\fancyrefasmslabelprefix}{Assumptions~#1}
\frefformat{vario}{\fancyreffiglabelprefix}{Figure~#1}
\frefformat{vario}{\fancyrefapplabelprefix}{Appendix~#1} 
\frefformat{vario}{\fancyrefproplabelprefix}{Proposition~#1}
\frefformat{vario}{\fancyrefalglabelprefix}{Algorithm~#1}
\frefformat{vario}{\fancyrefeqlabelprefix}{(#1)}
\frefformat{vario}{\fancyreftbllabelprefix}{Table~#1}
\frefformat{vario}{\fancyrefestilabelprefix}{Estimator~#1}

\newcommand{\nwidth}{n_w}

\newcommand{\dataset}{\mathbb{D}}
\newcommand{\searchspace}{\mathcal{G}}

\safemath{\dictab}{[\,\dicta\,\,\dictb\,]}

\safemath{\ysig}{\bmy}
\safemath{\ysighat}{\hat{\ysig}}
\safemath{\ysigdim}{M}
\safemath{\xsig}{\bmx}
\safemath{\xsigdim}{N}
\safemath{\nx}{n_x}
\safemath{\zsig}{\bmz}
\safemath{\zsigdim}{\ysigdim}
\safemath{\rsig}{\bmr}
\safemath{\Adict}{\bA}
\safemath{\Adicttilde}{\widetilde{\Adict}}
\safemath{\Adictdim}{\outputdim\times\xsigdim}
\safemath{\avec}{\bma}
\safemath{\avectilde}{\tilde{\avec}}
\safemath{\Bdict}{\bB}
\safemath{\Bdicttilde}{\widetilde{\Bdict}}
\safemath{\Cdict}{\bC}
\safemath{\cvec}{\bmc}
\safemath{\Ddict}{\bD}
\safemath{\Ddictdim}{\ysigdim\times\xsigdim}
\safemath{\dvec}{\bmd}
\safemath{\Ddicttilde}{\widetilde{\bD}}
\safemath{\Bonb}{\bB}
\safemath{\bvec}{\bmb}
\safemath{\Bonbdim}{\ysigdim\times\ysigdim}
\safemath{\noisedim}{\ysigim}
\safemath{\err}{\bme}
\safemath{\errdim}{\ysigdim}
\safemath{\errset}{\setE}
\safemath{\nerr}{n_e}
\safemath{\delop}{\bP_\errset}
\safemath{\delopc}{\bP_{{\errset}^c}}

\safemath{\cplxi}{\imath}
\safemath{\cplxj}{\jmath}

\safemath{\dict}{\matD}
\safemath{\inputdim}{N}		
\safemath{\outputdim}{M}		
\safemath{\sparsity}{S}	
\safemath{\inputdimA}{{N_a}}	
\safemath{\inputdimB}{{N_b}}	
\safemath{\elemA}{{n_a}}	
\safemath{\elemB}{{n_b}}	
\safemath{\resA}{\matR_a}	
\safemath{\resB}{\matR_b}	
\safemath{\subD}{\matS} 
\safemath{\subA}{\matS_a} 
\safemath{\subB}{\matS_b} 
\safemath{\dicta}{\matA} 	
\safemath{\dictb}{\matB} 	
\safemath{\hollowS}{H}
\safemath{\hollowA}{H_a}
\safemath{\hollowB}{H_b}
\safemath{\cross}{Z}
\safemath{\coh}{\mu_d}			
\safemath{\coha}{\mu_a}			
\safemath{\cohb}{\mu_b}			
\safemath{\mubs}{\nu}	
\safemath{\cohm}{\mu_m} 
\safemath{\dictset}{\setD}	
\safemath{\dictsetp}{\dictset(\coh,\coha,\cohb)}	
\safemath{\dictsetgen}{\dictset_\text{gen}}
\safemath{\dictsetgenp}{\dictsetgen(\coh)}
\safemath{\dictsetonb}{\dictset_\text{onb}}
\safemath{\dictsetonbp}{\dictsetonb(\coh)}

\safemath{\leftside}{U}
\safemath{\rightsideA}{R_a}
\safemath{\rightsideB}{R_b}

\safemath{\indexS}{\setI_S} 

\safemath{\na}{n_a}			
\safemath{\nb}{n_b}			
\safemath{\coeffa}{p_i}	
\safemath{\coeffb}{q_j}	
\safemath{\seta}{\setP}		
\safemath{\setb}{\setQ}     
\safemath{\setw}{\setW}	
\safemath{\setz}{\setZ}	
\safemath{\cola}{\veca}		
\safemath{\colb}{\vecb}		
\safemath{\cold}{\vecd}		
\safemath{\inputvec}{\vecx} 	
\safemath{\error}{\vece}	
\safemath{\noiseout}{\vecz} 	
\safemath{\inputvecel}{x}
\safemath{\inputveca}{\vecx_a}
\safemath{\inputvecb}{\vecx_b}
\safemath{\outputvec}{\vecy}	
\safemath{\lambdamin}{\lambda_{\mathrm{min}}}

\safemath{\elltwo}{\ell_2}
\safemath{\ellone}{\ell_1}
\safemath{\ellzero}{\ell_0}
\safemath{\ellinf}{\ell_\infty}
\safemath{\ellinftilde}{\ell_{\widetilde\infty}}
\safemath{\licard}{Z(\coh,\coha,\cohb)}
\safemath{\xsol}{\hat{x}}
\safemath{\xbord}{x_b}		
\safemath{\xstat}{x_s}		
\safemath{\xstatLone}{\tilde{x}_s}
\safemath{\scales}{\Theta} 
\safemath{\zeroes}{\mathbf{0}} 
\safemath{\thlone}{\kappa(\coh,\cohb)} 
\safemath{\constoneA}{\delta} 
\safemath{\constoneB}{\epsilon} 
\safemath{\nlarge}{L}				   
\safemath{\sumlarge}{S_\nlarge}
\safemath{\maxlarger}{P_\nlarge}	   
\safemath{\Pzero}{\textrm{P0}}	
\safemath{\Pone}{\textrm{P1}}
\safemath{\vecfir}{\vecw}			 
\safemath{\vecsec}{\vecz}
\safemath{\elvecfir}{w}              
\safemath{\elvecsec}{z}				 
\safemath{\nlargefir}{n}
\safemath{\normout}{\gamma}
\safemath{\auxfun}{h}
\safemath{\supp}{\textrm{supp}}

\safemath{\indexa}{\ell}
\safemath{\indexb}{r}
\safemath{\indexc}{i}
\safemath{\indexd}{j}

\safemath{\project}{P}
\newcommand{\fakeparagraph}[1]{\noindent {\bf #1 }}

\usepackage[capitalise,nameinlink]{cleveref}

\title{\LARGE \bf
Cost-Aware Diffusion Active Search
}

\author{Arundhati Banerjee$^{1}$ and Jeff Schneider$^{1}$
\thanks{$^{1}$School of Computer Science,
        Carnegie Mellon University, 5000 Forbes Avenue, Pittsburgh PA 15213
        {\tt\small arundhat@andrew.cmu.edu}}%
}

\begin{document}

\maketitle
\thispagestyle{empty}
\pagestyle{empty}

\begin{abstract}
%
Active search for recovering objects of interest through online, adaptive decision making with autonomous agents requires trading off exploration of unknown environments with exploitation of prior observations in the search space. 
Prior work 
has proposed 
information gain and Thompson sampling based myopic, greedy approaches for agents to actively decide query or search locations when the number of targets is unknown. 
Decision making algorithms in such partially observable environments have also shown that agents capable of lookahead over a finite horizon outperform myopic policies for active search. 
Unfortunately, lookahead algorithms typically rely on building a computationally expensive search tree that is simulated and updated based on the agent's observations and a model of the environment dynamics. 
Instead, in this work, we leverage the sequence modeling abilities of diffusion models to sample lookahead action sequences that balance the exploration-exploitation trade-off for active search without building an exhaustive search tree. 
We identify 
the optimism bias 
in prior diffusion based reinforcement learning approaches when applied to the active search setting and propose mitigating solutions for efficient cost-aware decision making with both single and multi-agent teams.  
Our proposed algorithm outperforms standard baselines in offline reinforcement learning in terms of full recovery rate and is computationally more efficient than tree search in cost-aware active decision making. 

\end{abstract}

\section{Introduction}
\label{sec:Intro}

Autonomous physical agents (robots) have great potential in practical applications like search and rescue operations over large unstructured regions, localization and mapping of unknown surroundings for environment monitoring, construction and manufacturing, etc. 
In such tasks, agents 
successfully achieve the desired goal over multiple timesteps by adapting their behavior or actions to their observations in the environment. 
\emph{Active search} describes such settings where agents make online adaptive decisions about interacting with their environment to identify objects of interest. 
For example, autonomous robots deployed in the aftermath of mining accidents \cite{5069840}, earthquakes \cite{murphy2011use} and wildfires \cite{8848961} may decide where (sensing location) and how (field of view) to sense their surroundings to rescue survivors from the disaster zones \cite{Murphy2008}. 
Prior work in informative path planning proposed information gain based objectives for sequential sensing given start and goal locations \cite{braun2015info, ma2017active, flaspohler2019information}. 
In order to avoid the limitations of such algorithms in multi-agent settings, \cite{kandasamy2019myopic, ghods2021decentralized} introduced the Thompson sampling framework for decentralized multi-agent active search with asynchronous inter-agent communication. 
%
Notably most of these prior approaches rely on myopic decision making which has been shown to be sub-optimal compared to lookahead planning, especially in environments which are not fully observable and where agents may encounter observation noise \cite{jiang2017efficient, jiang2019cost, fischer2020information}. 
\cite{banerjee2023cost} proposed a Monte Carlo Tree Search based sequential decision making algorithm called CAST which utilizes finite horizon lookahead for cost-awareness and outperforms myopic greedy approaches in multi-agent active search. 
%
%
Unfortunately, similar to most tree search based approaches to long horizon planning, CAST encounters scalability challenges as the dimension of the action space or search space increases. 
Another approach to decision making for active search proposes training reinforcement learning policies with appropriate belief space representations \cite{igoe2021multi}. 
With online data collection being expensive for these applications, the real world performance of such frameworks are determined by the quality of the offline training data, the availability of realistic simulators to overcome the simulation to reality gap, as well as the inference speed and robustness of the trained policies \cite{tang2025deep}.

In this work, we focus on addressing the challenges of lookahead decision making in active search by leveraging learning based approaches to amortize the cost of trajectory rollouts in the agent's search space. 
Specifically, we explore the ability of diffusion models to generate sequences of actions that can balance exploration with exploitation 
when trained with a noisy dataset collected by a myopic greedy agent covering few optimal action sequences in the search space. 
We identify that 
diffusion models may exhibit 
\emph{optimism bias} when used as generative planners in such 
problem formulations with stochastic optimal policies and observation noise. 
In contrast, our proposed approach called Cost-aware Diffusion Active Search (CDAS) addresses these challenges and leverages gradient-guided diffusion for cost-aware decision making outperforming baselines in myopic and lookahead active search in terms of both full recovery rate and inference time. 
%

\section{Problem Formulation}
\label{sec:genplan_setup}

\begin{figure}[htp]
    \centering
    \begin{subfigure}{0.45\linewidth}
    \centering
    \includegraphics[width=0.65\textwidth]
    {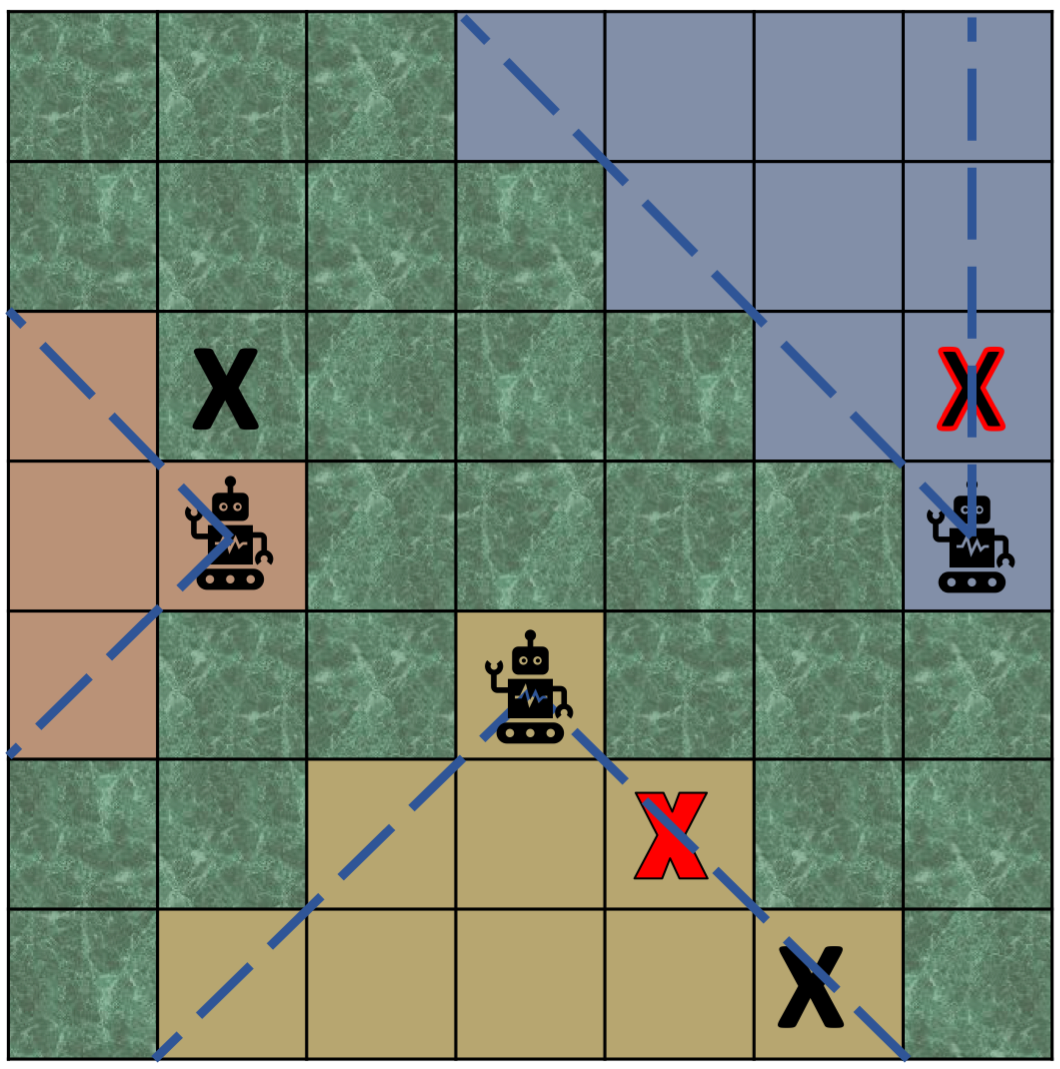}
    \caption{}
    \label{subfig:genplan-setup}
    \end{subfigure}%
    \begin{subfigure}{0.45\linewidth}
    \centering
    \includegraphics[width=0.95\textwidth]{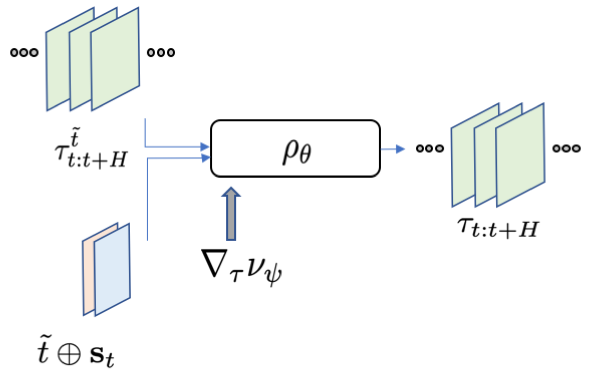}
    \caption{}
    \label{subfig:diffnet}
    \end{subfigure}
    \caption{\textbf{(a) Problem setup.} Agents sense different parts of the environment looking for OOIs. True OOIs are crossed in black. Targets detected by the agent in its field of view are crossed in red. \textbf{(b) Modeling for cost-aware diffusion in active search.} Agents sample lookahead action sequences of length $H$ conditioned on the current belief state and with gradient guidance from the estimated cumulative discounted return over the entire sequence.}
    \label{fig:explain}
\end{figure}

Consider a team of $\teamsize$ ground robots 
searching for an unknown number ($\numtargets$) of targets or objects of interest in a 2-dimensional (2D) region $\searchspace$ of length $n_\ell$ and width $\nwidth$ (\cref{subfig:genplan-setup}). 
We assume a gridded search environment 
$B \in \{0,1\}^{n_\ell \times n_w}$ which the agents recover through active search. 
$\bm\beta\in\{0,1\}^n$ is the flattened vector representation of $B$ and it is our search vector with $k$ non-zero entries at the $k$ target locations unknown to the agents. 
$n = n_\ell\times n_w$ is the total number of grid cells. $k << n$. 
We further assume that agents can self-localize in a global coordinate system. 

\subsection{Sensing model}
\label{subsec:genplan_sensingmodel}

$\bmx_t \in \{0,1\}^{Q_t\times n}$ is a region sensing action at time $t$ over $Q_t$ 
grid cells with 
a $90\degree$ field of view (FOV). 
Each row of this matrix $\bmx_t$ is a one-hot vector indicating the grid cell being sensed. 
%
%
We assume a linear sensing model with i.i.d white noise at each grid cell: 
\begin{equation}
    \bm{y}_t = \bmx_t \bm\beta + \bm\omega_t\label{eq:genplan_sensing}
\end{equation}
where $\bm\omega_t \sim \mathcal{N}(\mathbf{0},\sigma^2\mathbf{1}_{Q\times 1})$, $\sigma^2$ is the (constant) variance of observation noise.

\subsection{Cost model}
\label{subsec:genplan_cost}

We consider realistic travel and sensing costs for agents' actions. 
An agent travelling from sensing location $x_t$ (for sensing action $\bmx_t$) to $x_{t+1}$ (for sensing action $\bmx_{t+1}$) incurs a travel time cost $c_d (x_t, x_{t+1})$ when the Euclidean distance $d(x_t, x_{t+1})$ is traveled at a constant speed $v$. 
Separately, executing the sensing action $\bmx_{t+1}$ at location $x_{t+1}$ incurs a time cost $c_s(x_{t+1})$. 
Therefore, $T$ time steps after starting from location $x_0$, an agent $j$ has executed actions $\{\bmx_t^j\}_{t=1}^{T}$ and incurs a total cost 
$C^j(T) = \sum_{t=1}^{T} (c_{d}(x_{t-1}^j,x_{t}^j) + c_{s}(x_t^j))$.

\subsection{Communication}
\label{subsec:genplan_comm}

We assume communication is unreliable 
and 
there is no central controller to coordinate communication and/or actions among the team. 
Instead, agents 
communicate \emph{asynchronously} when possible, without having to wait for others to complete their tasks, and thus leverage information sharing in a decentralized multi-agent team \cite{ghods2021decentralized}.

To recover the search vector $\bm\beta$ by actively identifying all the targets, at each time $t$, an agent $j$ chooses the best sensing action $\bmx_t^j$ based on its belief about the environment given the measurements available thus far in its measurement set $\bD_t^j$. 
%
For a single agent, our problem reduces to sequential decision making with the measurement set $\bD_t^1 = \{(\bmx_1,\bmy_1),\dots,(\bmx_{t-1},\bmy_{t-1})\}$ available to the agent at time $t$. 
Note that the set of available past measurements need not remain consistent across agents due to communication unreliability.
In the general multi-agent setting, following our communication constraints, our goal is to learn a non-myopic policy for decentralized and asynchronous (multi-agent) active search so that agents recover $\bm\beta$ with as few measurements $T$ as possible while also optimizing for the total incurred cost $\sum_j C^j(T)$.

\section{Related Work}
\label{sec:genplan_relatedwork}

Diffusion models \cite{songscore} have recently emerged as a popular approach to generative modeling, particularly for image and video data. 
Diffusion models provide a framework to generate samples from an unknown distribution $p^*$ by iteratively transforming samples from a simpler (i.e. easy to sample from) distribution, for example, a standard Gaussian distribution.\footnote{For a thorough review of this topic, please refer to \cite{songdenoising, songscore, nakkiran2024step}.}
Given samples $x$ from the training data distribution, the forward noising step first adds Gaussian noise sequentially with a fixed noise schedule over $T_{\text{diff}}$ steps and the reverse denoising step then trains the generative model to recover 
$x$ by removing this added noise. 
Finally we can sample from such a trained model via Langevin dynamics \cite{ho2020denoising}.  
%
%
During the reverse denoising process, the generated samples can be biased or guided towards the support of the desired distribution with two main approaches. 
%
%
In classifier-free guidance \cite{dhariwal2021diffusion}, a conditional diffusion model is trained to generate samples given a particular value of the conditioning variable. 
In contrast, classifier guidance \cite{ho2022classifier} uses the gradient from a separate model $\nabla\log \nu(c|x)$ to bias the sample generation process 
during inference. 

Recently, \cite{janner2022planning, ajayconditional} consider 
offline RL as a sequence modeling problem 
and train diffusion models for agents to generate a sequence of observed states and state-conditioned actions to reach their goal. 
\cite{janner2022planning} introduced diffusion models with reward-gradient guidance as planners in offline RL whereas \cite{alonso2024diffusion} proposed conditional diffusion based world models for training RL policies. 
While these approaches primarily leverage the representation capacity of diffusion models for generating multi-modal distributions over feasible states, \cite{zhou2024diffusion} models diffusion jointly over \emph{both} states and actions in a reward-gradient guided MPC-style lookahead diffusion algorithm and outperforms standard RL baselines. 
%
%
%
%
%
But \cite{zhou2024diffusion} only considers environments with deterministic state transitions 
and access to low dimensional state information. 
As we discuss later, stochasticity in state transition due to observation noise and unknown ground truth pose a significant challenge in the application of diffusion models for decision making in active search. 
%
%
%


%
Following the centralized training with decentralized execution (CTDE) framework in cooperative multi-agent settings, \cite{shaoul2024multi} combined single-agent diffusion models with constraint-based planners for multi-agent path finding whereas \cite{zhu2023madiff} introduced a cross-attention layer in the network for modeling inter-agent coordination in state prediction. 
%
In both cases, 
a central controller would be required to coordinate the trajectories of the different agents. 
In contrast, there has been little focus on the asynchronous decentralized multi-agent setting which is of interest in this work. 



Next, we show that extrapolating the success of diffusion modeling to active search is non-trivial and focus on some of the challenges we observed and possible approaches to mitigating them for lookahead decision making in multi-agent active search.

\section{Diffusion models for active search} 
 \label{sec:genplan_diffusion}


%

In our problem formulation (\cref{sec:genplan_setup}), 
agents recover an unknown search vector $\bm\beta$ using noisy observations from region sensing actions. 
The full recovery rate is measured by the number of targets that were correctly identified by all the agents over $\horizon$ steps. 
In this setting, depending on the unknown ground truth $\bm\beta^*$ for a particular run, the same sequence of actions could lead to different noisy observations and receive a high or low full recovery reward. 
Moreover, the agent's posterior belief about the state space depends on both the sequence of actions and the resulting observations. 
%
Therefore training a generative model to learn a distribution over state-action sequences in active search is quite challenging and it involves dealing with non-negligible stochasticity and multi-modality in the training data. 
Further, 
agents do not have fixed start and goal locations. Instead agents must trade-off exploration and exploitation over the search space and adapt their subsequent behavior to prior observations. 
This leads to optimism bias with diffusion models that generate joint state-action sequences, further validating prior observations 
using transformers 
in similar stochastic POMDP formulations \cite{villaflor2022addressing}. 
%


\subsection{Belief representation}

%
Each agent's belief over the search vector $\bm\beta$ is initialized with a Gaussian prior 
%
%
$p_0(\bm\beta) = \normal(\hat{\bm\beta}_0, \hat{\bm\Sigma}_0)$ 
where the prior mean vector $\hat{\bm\beta}_0 = \frac{1}{n}\mathbf{1}_{n\times 1}$ and the prior covariance matrix $\hat{\bm\Sigma}_0 = \sigma^2\mathbf{I}_{n\times n}$. 
%
%
%
Since agents follow a linear sensing model (\cref{eq:genplan_sensing}), 
we can use a Kalman filter \cite{kalman1960new} to update the posterior belief $p_t(\bm\beta) = \normal(\hat{\bm\beta}_t, \hat{\bm\Sigma}_t)$ over the search space 
%
using 
alternate prediction and update steps. 
$\itime$ denotes the decision making time step.\footnote{Note that we have to carefully distinguish the diffusion timestep which we denote using superscript from the decision making timestep which we denote using subscript.} 

In order to capture the spatial grid representation with diffusion, we represent the agent's belief state by concatenating the posterior mean and variance matrices along the channel dimension, i.e. $\bms_t = \hat{\bm\beta}_t \oplus \hat{\bm\Sigma}_t \in [0,1]^{2\times n_\ell \times n_w}$. 
Since our sensing model assumes i.i.d observation noise among the grid cells in its field of view, we will only account for the diagonal of $\hat{\bm\Sigma}_t$ in $\bms_t$. 


\subsection{Lookahead plan generation with diffusion models}
\label{subsec:genplan_method_details}

\begin{figure}
    \centering
    \includegraphics[width=0.5\linewidth]{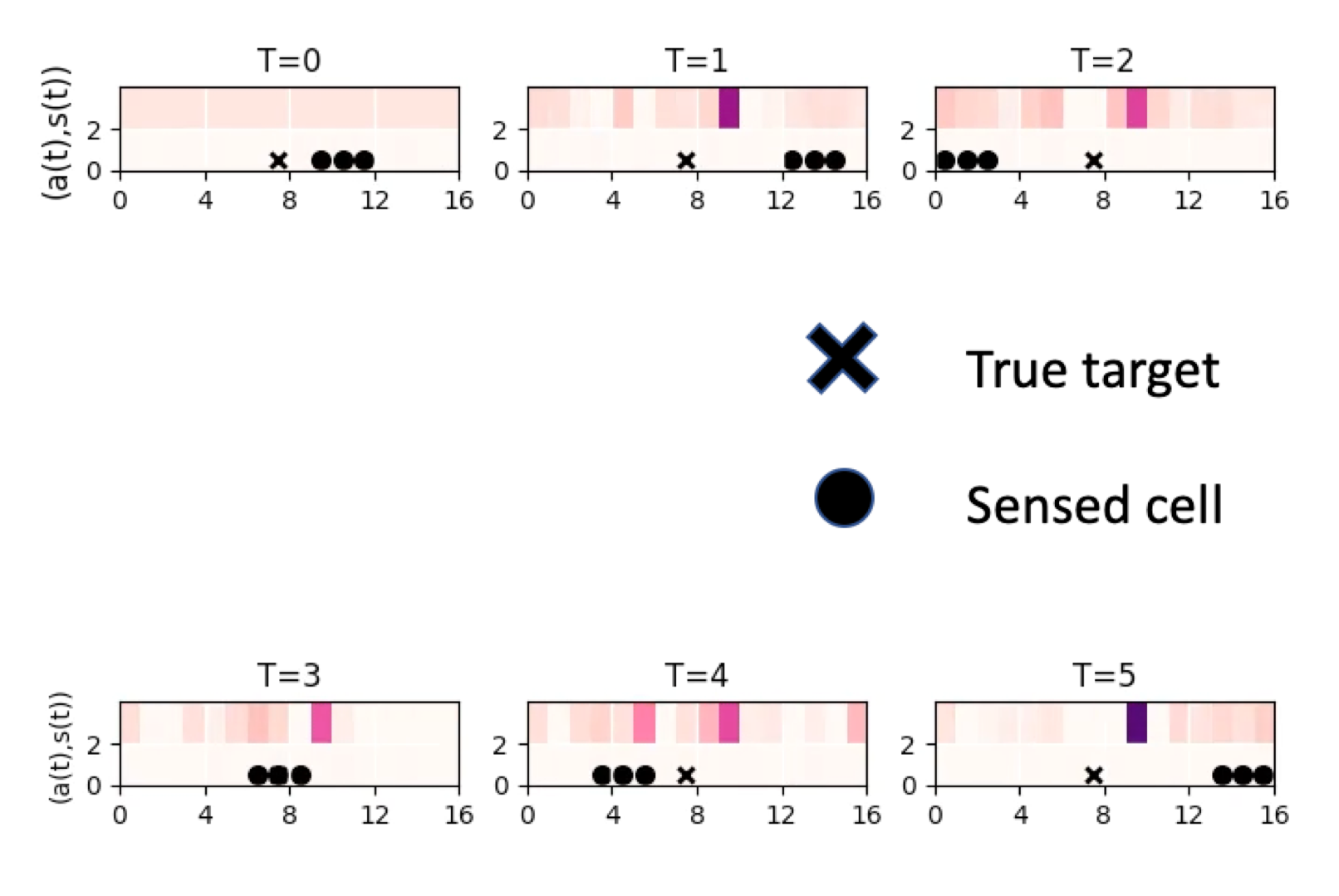}
    \caption{\textbf{Optimism bias in samples generated by diffuser \cite{janner2022planning} for active search.} Top row in each block (for $T=0\dots 5$) is the posterior mean estimate 
    (darker shade implies higher value) 
    and bottom row indicates true target location and the FOV of the sensing action. Diffusion modeling over joint state-action space with gradient guided sampling of state-action sequences learns to generate deterministic target detection after the first generated action. 
    In other words, generated samples are biased to assume the target will be detected in the first timestep. 
    Such sampled action sequences do not balance exploration-exploitation, explaining why diffuser is unable to generate optimal coverage action sequences for active search. 
    }
    \label{fig:optimism_bias}
\end{figure}

%
%
%
%
Prior work 
\cite{janner2022planning, chi2024diffusionpolicy} attempts to fold as much of the planning stage as possible into the generative modeling framework so that sequential decision making becomes equivalent to drawing samples from a trained diffusion model. 
In our problem setting, we represent a state-action sequence over a lookahead horizon $H$ as a sequence of 3-channel images, where each image in the sequence is a concatenation of $\bms_t$ and $\bmx_t$ along the channel dimension. 
Following \cite{janner2022planning}, a diffusion model trained to generate samples $\{\bms_{t'},\bmx_{t'}\}_{t'=t}^{t+H}$ 
would lead to $\bmx_{t:t+H}$ as the generated action plan chosen for the subsequent timesteps at $\itime$. 
Unfortunately, as we further illustrate in \cref{fig:optimism_bias}, this formulation of the generative model leads to
sampling of overly optimistic state-action sequences 
where the actions do not balance exploration-exploitation in the search region. 
We refer to this as the \emph{optimism bias} for diffusion based planning in active search. 

To 
address this, we instead train a diffusion model to generate a lookahead plan as a sequence of actions $\bmx_{t:t+H}$ conditioned on the agent's belief state $\bms_t$ that maximizes an objective $J(\bms_t, \bm\tau_t)$. 
Hence 
we model lookahead decision making as denoising diffusion sampling 
a sequence of single-channel images, each image $\bmx_{\itime'} \in \{0,1\}^{n_\ell\times n_w},\;\itime'\in\{\itime,\dots,\itime+H\}$ 
representing a region sensing action. 
%

\fakeparagraph{Training loss: }
%
%
In order to recover targets with as few measurements $T$ as possible, 
sampled action sequences from the diffusion model 
should maximize the expected 
full recovery reward given the agent's belief state. 
This conditional sampling framework is captured by gradient guidance during denoising diffusion sampling \cite{janner2022planning}. 
So we train both a trajectory generation model $\rho_\theta$ and a 
return estimation model $\nu_\psi$ by optimizing the following loss functions. 

%
%

\begin{align}
    \quad\mathcal{L}(\theta) = \mathbf{E}_{\substack{
        \tilde{\itime}\sim U(0,T_{\text{diffusion}})\\
        \bms,\bm\tau\sim \dataset_M}
    }[\|\bm\tau - \rho_\theta(\bms, \bm\tau^{\tilde{\itime}}, \tilde{\itime})\|_2^2]\label{eq:diffusion_traj_loss}
\end{align}

\begin{align}
    \quad\mathcal{L}(\psi) = \mathbf{E}_{\substack{
        \tilde{\itime}\sim U(0,T_{\text{diffusion}})\\
        \bms,\bm\tau,\bmr_{\bm\tau} \sim \dataset_M}}[\|R_\tau^\itime - \nu_{\psi}(\bms,\bm\tau^{\tilde{\itime}},\tilde{\itime})\|_2^2]\label{eq:diffusion_return_loss}
\end{align}

$\tilde{\itime}$ is the forward noising step, $U(\cdot,\cdot)$ denotes uniform distribution and $\mathbb{D}_M$ is the training dataset. 
We empirically observed that the optimization in \cref{eq:diffusion_return_loss} is more stable without forward noising ($T_{\text{diffusion}}=0$), especially with increasing search space dimensions. 
At any step in the sequential decision making process, our simulation assigns the agent 
a reward of $+1$ for fully recovering the target(s), otherwise a penalty $-1$. 
In order to ensure that gradient guidance from $\nu_\psi$ balances exploration with exploitation in the sampled $\tilde{\bm\tau}\sim\rho_\theta$, the episodes in $\mathbb{D}_M$ are labeled with the expected per-timestep reward for training $\nu_\psi$. 
We define $R_\tau^\itime = \sum_{\itime'=0}^{H-1}\gamma^{\itime'}\bmr_{\bm\tau}[\itime+\itime']$ with $r = \mathbf{r}_{\bm\tau}[\itime]$ given by 
\begin{equation}
    r = \mathbb{E}_{\bmy|\action,\tilde{\bm\beta}} \mathbb{E}_{\tilde{\bm\beta}} [\mathbf{1}\{\tilde{\bm\beta}=\hat{\bm\beta}'\} - \mathbf{1}\{\tilde{\bm\beta}\not=\hat{\bm\beta}'\}]\label{eq:genplan_onestep_reward}
\end{equation}
where $\tilde{\bm\beta}\sim p_\itime(\bm\beta)$, $\bmy$ is the noisy observation from a sensing action $\bmx$ (\cref{eq:genplan_sensing}) and $\hat{\bm\beta}'$ is the resulting posterior estimated mean of the agent's belief.

%



\fakeparagraph{Data generation: }
%
%
%
The training dataset $\mathbb{D}_M$ consists of 
$M$ episodes simulated 
with an information-greedy active search agent (\cref{eq:genplan_IG_action}) over $T$ timesteps. 
Note the agent's behavior policy is different from our definition of $R_{\tau}$ above. 
%
The 
search vector $\bm\beta$ differs across episodes 
and we set $\numtargets=1$. 
For a lookahead horizon $H$, the training samples are obtained by 
chunking each episode into sequences of length $H$ given by 
$(\bms_{\itime:\itime+H}, \action_{\itime:\itime+H}, r_{\itime:\itime+H})$ across 
$\itime \in \{0,\dots,\horizon-H\}$. 

\subsection{Inductive biases in RL for active search}

The performance of generative models 
in different applications like language modeling or video generation 
often depends on the network architecture inducing the appropriate inductive bias for the machine learning task. 
%
Prior work has explored equivariant representation learning in model-free RL \cite{mondal2022eqr} and training Q-learning or actor-critic algorithms with equivariant network architectures \cite{nguyen2023equivariant, wangmathrm} to improve the sample efficiency and generalization performance in different tasks. 
Notably in active search with region sensing actions and information gain based rewards, the expected return (i.e. cumulative information gain) over a horizon $H$ given a 
belief state 
is permutation invariant to the sequence of actions. 
Further the posterior belief update over the grid cells in the search space is equivariant with the sensing action's FOV. 
Such symmetries are captured by appropriately defined convolution layers and also in graph neural networks, motivating our exploration of the following neural network architectures for $\rho_\theta$ and $\nu_\psi$.\footnote{We will provide a publicly available code repository upon publication.}

First, we consider a U-Net architecture \cite{ronneberger2015u} with 2D convolution layers which takes as input the action sequence $\bm\tau \in \{0,1\}^{H \times n_\ell \times n_w}$ (similar to the sequence of frames input in image based diffusion). 
As \cref{subfig:diffnet} shows, we additionally use FiLM conditioning \cite{perez2018film} of the concatenated state $\bms$ and diffusion timestep $\tilde{\itime}$ embedding as input to train a conditional diffusion model. 

    Second, we also consider a graph attention based diffusion framework. 
    Consider an undirected graph $\mathcal{G}$ where each node corresponds to a grid cell in the discretized search space. 
    $\mathcal{G}$ is fully connected with edge weights $e_{ij}$ proportional to the Manhattan distance between grid cell $i$ and $j$. 
    Each node $i$ is represented by a 
    vector $$\bmf_i = [{\bm\tau}_{\itime:\itime+H, i}^{\tilde{\itime}}\;\hat{\bm\beta}_{\itime,i}\; \hat{\bm\Sigma}_{\itime,i}\;\tilde{\itime}_{\text{emb}}]_{1\times (H+3)}.$$ 
    Here $\hat{\bm\beta}_{\itime,i}$ and $\hat{\bm\Sigma}_{\itime,i}$ indicate the $i^{th}$ index element of the 
    posterior mean and variance. 
    ${\bm\tau}^0_{\itime:\itime+H,i}$ is a vector of $0$s and $1$s indicating whether the $i^{th}$ grid cell is sensed ($1$) or not ($0$) for each of the $H$ 
    timesteps in $\bm\tau$. 
    The superscript $0$ indicates zero added diffusion noise. 
    $\tilde{\itime}$ indicates the diffusion (de)noising timestep. 
    $\tilde{\itime}_{\text{emb}}$ is a (learnable) embedding of the diffusion timestep. 
    %
    %
    In this representation, the temporal dimension of the sequence modeling problem is captured by the node features and the spatial geometry by the node adjacency matrix. 
    Following prior work \cite{velivckovic2018graph} which proposed Graph Attention Networks (GAT), 
    %
    we also use graph attention layers in our network, where the edge weights are used to learn a soft attention mask. 
    %
    
    %
    %
    %

In the trajectory generation network $\rho_\theta$, the 
output 
is a sequence of actions over $H$ steps 
%
whereas 
$\nu_\psi$ 
outputs a scalar value for the estimated return. 
We separately train $\rho_\theta$ and $\nu_\psi$ with the training dataset $\dataset_M$ 
and follow the gradient guided sampling framework similar to \cite{janner2022planning} to incorporate cost-awareness (\cref{line:genplan_algo7l5} and  \cref{line:genplan_algo7l6}). 
Therefore we additionally train a distance estimator $d_\varphi$ with the same graph neural network architecture described above, but using only the sequence of actions $\bm\tau$ as input and the cumulative euclidean distance traveled by the agent as output. 
\cref{algo:genplan_gradguidance} outlines the inference algorithm for active search with diffusion. 
$\bm\Sigma_{\text{diff}}^{\tilde{t}}$ is the specified noise schedule for diffusion model training. 


%

\begin{algorithm}[htp]
    \caption{Cost-Aware Diffusion active search (CD-AS)}
    \begin{algorithmic}[1]
    \State{\textbf{Input:} $\rho_\theta$, $\nu_\psi$, $d_\varphi$, $\bms_\itime, \text{current location } x_{\itime-1}$} 
    \For{each of $N_{\text{diff}}$ samples}\label{line:genplan_algobatchsize}
    \State{Initialize $\bm\tau_\itime^{T_{\text{diff}}} \sim \normal(\bm0, \bm\identity)$}
    \For{$\tilde{\itime} = T_{\text{diff}}-1,\dots,0$}
    \State{$\bm\tau_\itime^{\tilde{\itime}-1} \sim \normal(\rho_\theta(\bms_\itime, \bm\tau^{\tilde{\itime}}, \tilde{\itime}) + \alpha\bm\Sigma_{\text{diff}}^{\tilde{\itime}}(\nabla\nu_\psi(\bms_\itime, \bm\tau_\itime^{\tilde{\itime}}, \tilde{\itime}) - \lambda\nabla d_\varphi(\bm\tau_\itime^{\tilde{\itime}}, \tilde{\itime})), \bm\Sigma_{\text{diff}}^{\tilde{\itime}})$}\label{line:genplan_algogradguidance}
    \EndFor
    \State{Compute $d_{\bm\tau_\itime^0}$: total ground truth distance traveled if $\bm\tau_\itime^0$ were executed starting from $x_{\itime-1}$}\label{line:genplan_algo7l5}
    \EndFor
    \State{Select $\bm\tau_\itime^0$ that maximizes $\nu_\psi(\bms_\itime, \bm\tau_\itime^0, 0) - \lambda\times d_{\bm\tau_\itime^0}$}
    \label{line:genplan_algo7l6}
    \State{Execute the first action $\bmx_\itime$ in $\bm\tau_\itime^0$. Observe $\bmy_\itime$.}
    \State{Update $\bms_{\itime+1}$ }
    \end{algorithmic}
\label{algo:genplan_gradguidance}
\end{algorithm}



\begin{algorithm}[htp]
    \caption{Multi-agent active search with diffusion}
    \begin{algorithmic}[1]
    \State{\textbf{Input:} $\rho_\theta$, $\nu_\psi$, $\bms_\itime^\iagent=(\hat{\bm\beta}_\itime^\iagent, \hat{\bm\Sigma}_\itime^\iagent), \text{current location } x_{\itime-1}^\iagent$} 
    \For{any available agent $j$}
    \State{Sample $\bm\tau_{\itime,\iagent}^0$ following \cref{algo:genplan_gradguidance}}
    \State{Execute the first action $\bmx_{\itime,\iagent}$ in $\bm\tau_{\itime,\iagent}^0$. Observe $\bmy_{\itime,\iagent}$.}
    \State{Share $(\bmx_{\itime,\iagent}, \bmy_{\itime,\iagent})$ with team.}
    \State{Update $\bms_{\itime+1,\iagent}$ using $(\bmx_{\itime,\iagent}, \bmy_{\itime,\iagent})$ and measurements received from other agents.}
    \EndFor
    \end{algorithmic}
\label{algo:genplan_gradguidance_multiagent}
\end{algorithm}

\cref{algo:genplan_gradguidance_multiagent} outlines our approach to decentralized and multi-agent cost-aware active search using the trained models. 
Note that 
$\mathbb{D}_M$ was collected in a single-agent setting, therefore we design each agent to draw its own samples independently 
conditioned on its current state. 
In this work, we do not focus on developing multi-agent coordination with diffusion, and it is left for 
future research.

\section{Experiments}
\label{sec:genplan_experiments}

We now describe the simulation setup for our experimental results. 
There are $J$ agents in a discretized search space 
tasked with recovering the unknown search vector $\bm\beta$, which is 
generated across different episodes using a uniform sparse prior with $k$ non-zero entries. 
Note that the agents are not aware of the number of targets (OOIs) $k$. 
We assume that agents are positioned at the centre of the grid cells they occupy and may 
move in any direction in the search space. 
%
%
Our performance metric is the full recovery rate, as previously defined. 
The plots show mean values with shaded regions indicating standard error over multiple trials, each trial differing only in the instantiation of the true position of the $k$ OOIs in $\bm\beta$. 
For simplicity, we evaluate with $k=1$ unless specified otherwise. 

\fakeparagraph{Baselines. }
We will use the following algorithms as baselines to compare both cost-awareness and full recovery rate. 

\par\emph{Myopic information-greedy active search} (EIG) \cite{ma2017active}:  
    Each agent selects an action to maximize the expected information gain or 
    the reduction in entropy of the posterior belief distribution following the action. 
    \begin{equation}
        \action_\itime = \argmax_\action \mathbb{E}_{\bmy}[H(p(\hat{\bm\beta})) - H(p(\hat{\bm\beta}|\action,\bmy))]\label{eq:genplan_IG_action}
    \end{equation}

\par\emph{Myopic Thompson Sampling} (TS) \cite{ghods2021decentralized}: Each agent selects an action to maximize the expected one-step lookahead reward assuming a Thompson sample $\tilde{\bm\beta}_\itime \sim p_\itime(\bm\beta)$ from the posterior belief is the unknown search vector. 
    \begin{equation}
        \action_\itime = \argmax_\action \mathbb{E}_{\bmy|\action,\tilde{\bm\beta}_\itime} [-\|\tilde{\bm\beta}_\itime - (\hat{\bm\beta}_{\itime+1}|\action,\bmy)\|_2^2]
    \end{equation}
    $\hat{\bm\beta}_{\itime+1}|\action,\bmy$ is the one-step lookahead posterior mean if action $\action$ were executed with observation $\bmy$.

\par\emph{Implicit Q-Learning} (IQL) \cite{kostrikovoffline} 
    is an offline RL algorithm that decouples policy improvement from value function estimation 
    thereby avoiding bootstrapping errors in the temporal difference loss. 
    IQL outperforms other approaches in offline RL particularly for tasks that require multi-step dynamic programming updates to extract optimal behavior by stitching together sub-optimal training data. 
    %
    Note that unlike diffusion-based approaches, IQL is not primarily focused on multi-modality in the dataset or stochastic policies. 

\par\emph{CAST} \cite{banerjee2023cost} is a Thompson sampling and pareto-optimization guided Monte Carlo Tree Search based algorithm for multi-agent cost-aware active search. 
We first adapt the implementation of CAST with our sensing model in \cref{eq:genplan_sensing}. 
    %
    CAST is an appropriate baseline for comparison with our proposed diffusion modeling based generative approach to amortized lookahead decision making. 
    Additionally, we will also compare the performance of cost-aware active search in CAST, with the inference-time cost-awareness of gradient-guided diffusion active search in CD-AS. 

\par\emph{Diffusion policy} \cite{chi2024diffusionpolicy} 
    proposes representing a robot’s visuomotor policy as a conditional denoising diffusion process and demonstrates advantages of using diffusion in learning multimodal policies by behavior cloning. 
    In contrast with our approach, this framework does not involve a reward model or gradient guidance to generate constraint-aware plans. 
    We use the diffusion policy implementation from 
    \cite{cadene2024lerobot} to compare against our diffusion based approach to active search.


\subsection{CD-AS in one-dimensional search space}

We first consider a one-dimensional (1D) search space discretized into $n = 16$ grid cells, with 
%
%
a single agent ($J=1$) and a single target ($k=1$). 
%
Each sensing action has a field of view of at most 3 grid cells in front of the agent along its viewing direction (east or west).   
We consider two observation noise settings $\sigma=\frac{1}{16}$ and $\sigma=0.2$ for both training and evaluation. 

During training, our diffusion based approach uses $H=8$ length sequences and the U-Net architecture for both $\rho_\theta$ and $\nu_\psi$, with $T_{\text{diff}} = 64$ denoising diffusion steps. 
At evaluation, we use a gradient guidance coefficient of $\alpha=10$. 
For each decision step, we sample $N_{\text{diff}}=10000$ action sequences of length $H=8$ conditioned on the current belief state, execute the first sensing action from this sequence and replan with the updated posterior belief at the next timestep. 

\begin{figure}[htp]
  \centering
    \begin{subfigure}{0.5\linewidth}
    \centering
    \includegraphics[width=\textwidth]{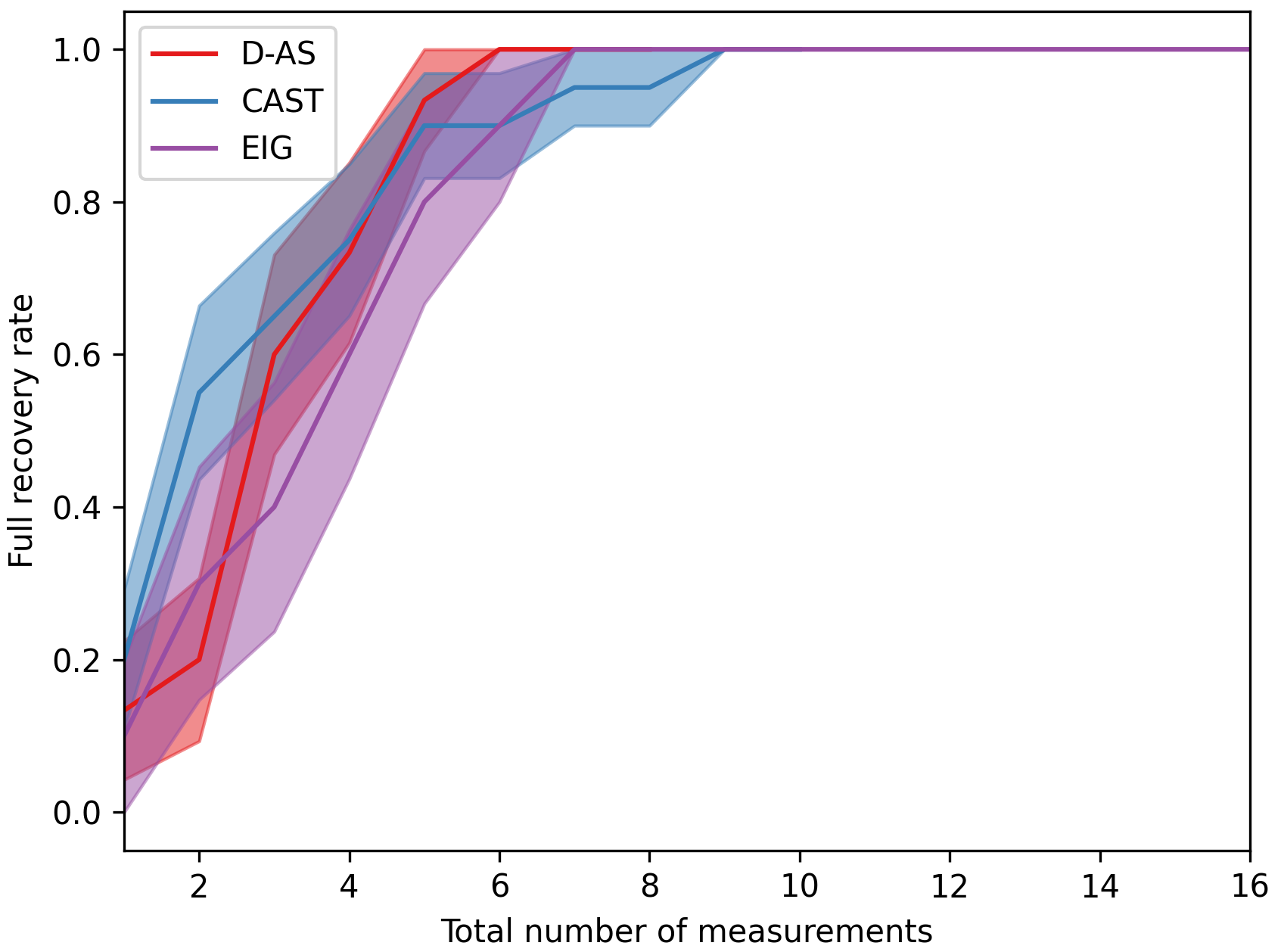}
    \caption{$\sigma=\frac{1}{16}$}
    \label{fig:1x16J1k1sigma1by16myopicfrec}
    \end{subfigure}%
    \begin{subfigure}{0.5\linewidth}
    \centering
    \includegraphics[width=\textwidth]{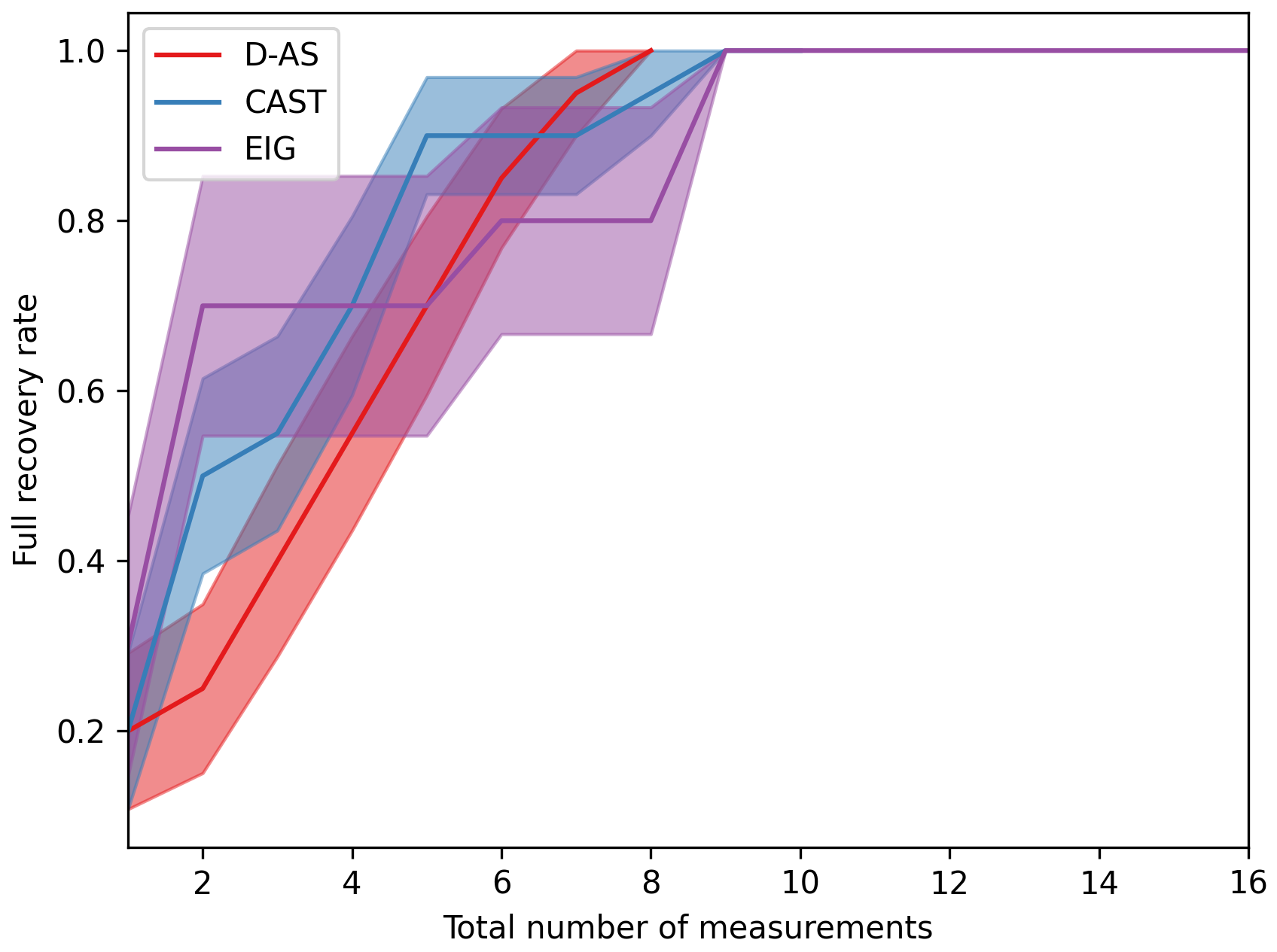}
    \caption{$\sigma=0.2$}
    \label{fig:1x16J1k1sigma0p2myyopicfrec}
    \end{subfigure}
    \caption{\textbf{Lookahead vs myopic decision making in 1D search space of size $n = 16$. $J=1$ agent. $k=1$ target. Low ($\sigma=\frac{1}{16}$) and high ($\sigma=0.2$) observation noise. }%
    Our diffusion based approach (D-AS) recovers the optimal sequence of actions and achieves full recovery faster than the myopic and shallow search tree based baselines. Plots show mean and standard error over 20 trials.} 
    \label{fig:genplanresults_1x16J1k1_myopicvslookahead}
\end{figure}

\cref{fig:genplanresults_1x16J1k1_myopicvslookahead} shows that our diffusion based lookahead decision making in CD-AS outperforms myopic (EIG) and shallow lookahead (CAST) baselines. 
%
%
CAST is simulated with a search tree of depth 2 and 5000 simulation episodes. 
In both 
the observation noise settings, diffusion is able to recover the target with minimum (optimal) number of measurements. 
%
Note that for a one dimensional grid of size 16, for the previously defined action space, a sequence of 6 non-overlapping sensing actions can cover the entire search space and in a low observation noise setting, all targets can be recovered with $H_{\text{opt}, \sigma\approx0}=6$. With higher observation noise, $H \geq H_{\text{opt},\sigma\approx0}$. 

\begin{figure}[htp]
  \centering
    \begin{subfigure}{0.45\linewidth}
    \centering
    \includegraphics[width=\textwidth]{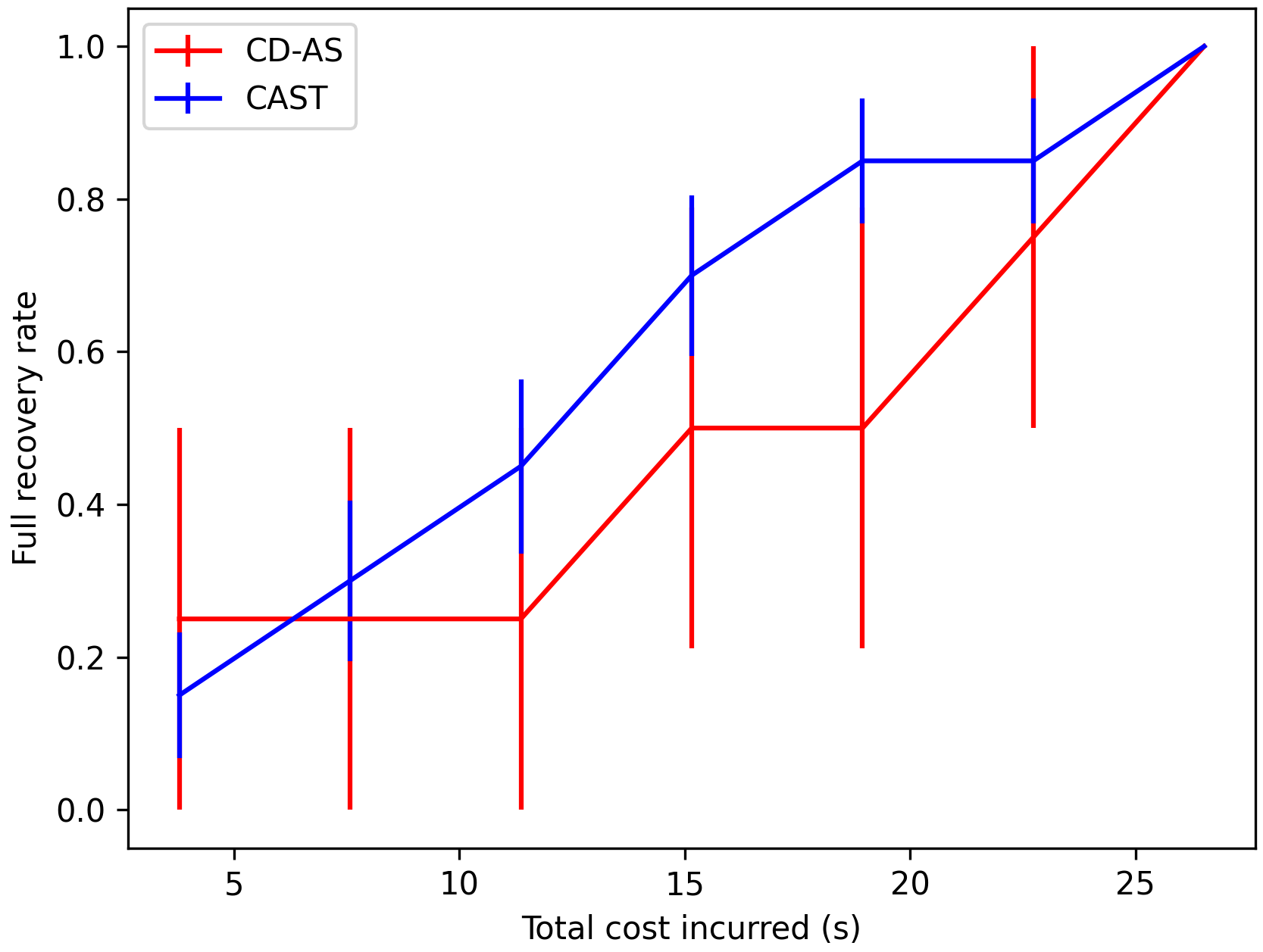}
    \caption{$\sigma=\frac{1}{16}\; c_s=0$s}
    \label{fig:1x16J1k1sigma1by16s0}
    \end{subfigure}%
    \begin{subfigure}{0.45\linewidth}
    \centering
    \includegraphics[width=\textwidth]{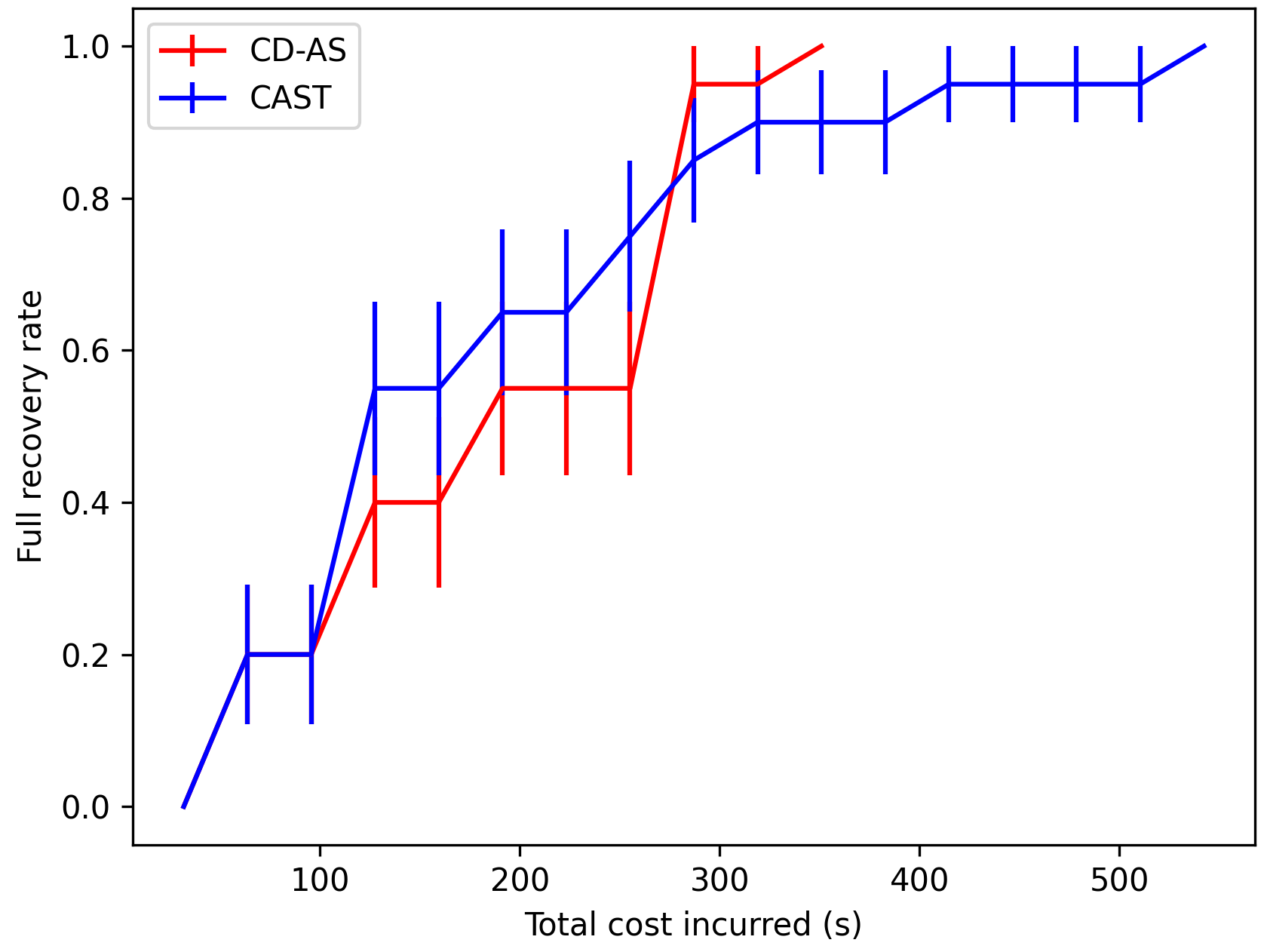}
    \caption{$\sigma=\frac{1}{16}\; c_s=50$s}
    \label{fig:1x16J1k1sigma1by16s50}
    \end{subfigure}
    \begin{subfigure}{0.45\linewidth}
    \centering
    \includegraphics[width=\textwidth]{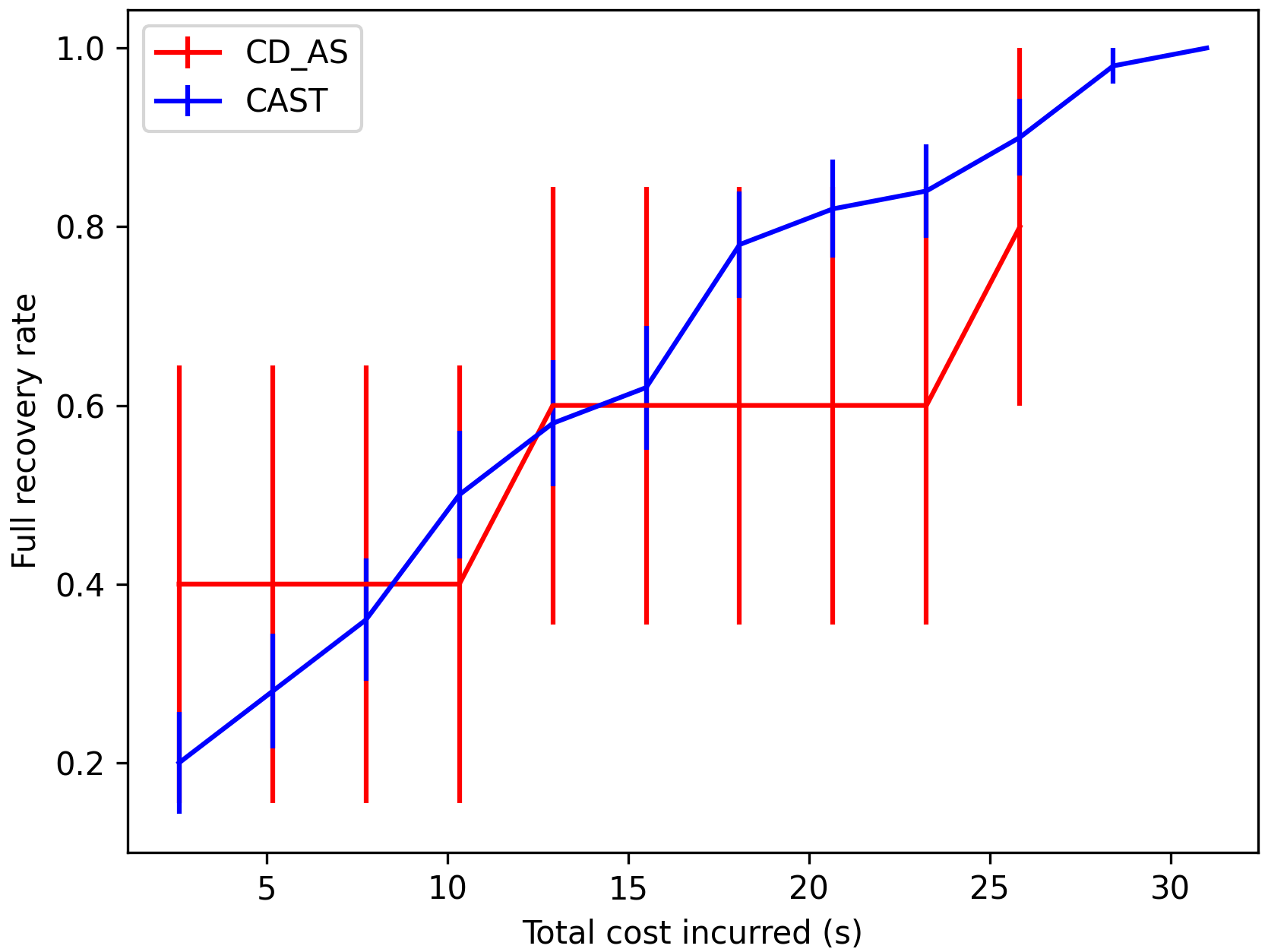}
    \caption{$\sigma=0.2\; c_s=0$s}
    \label{fig:1x16J1k1sigma0p2s0}
    \end{subfigure}%
    \begin{subfigure}{0.45\linewidth}
    \centering
    \includegraphics[width=\textwidth]{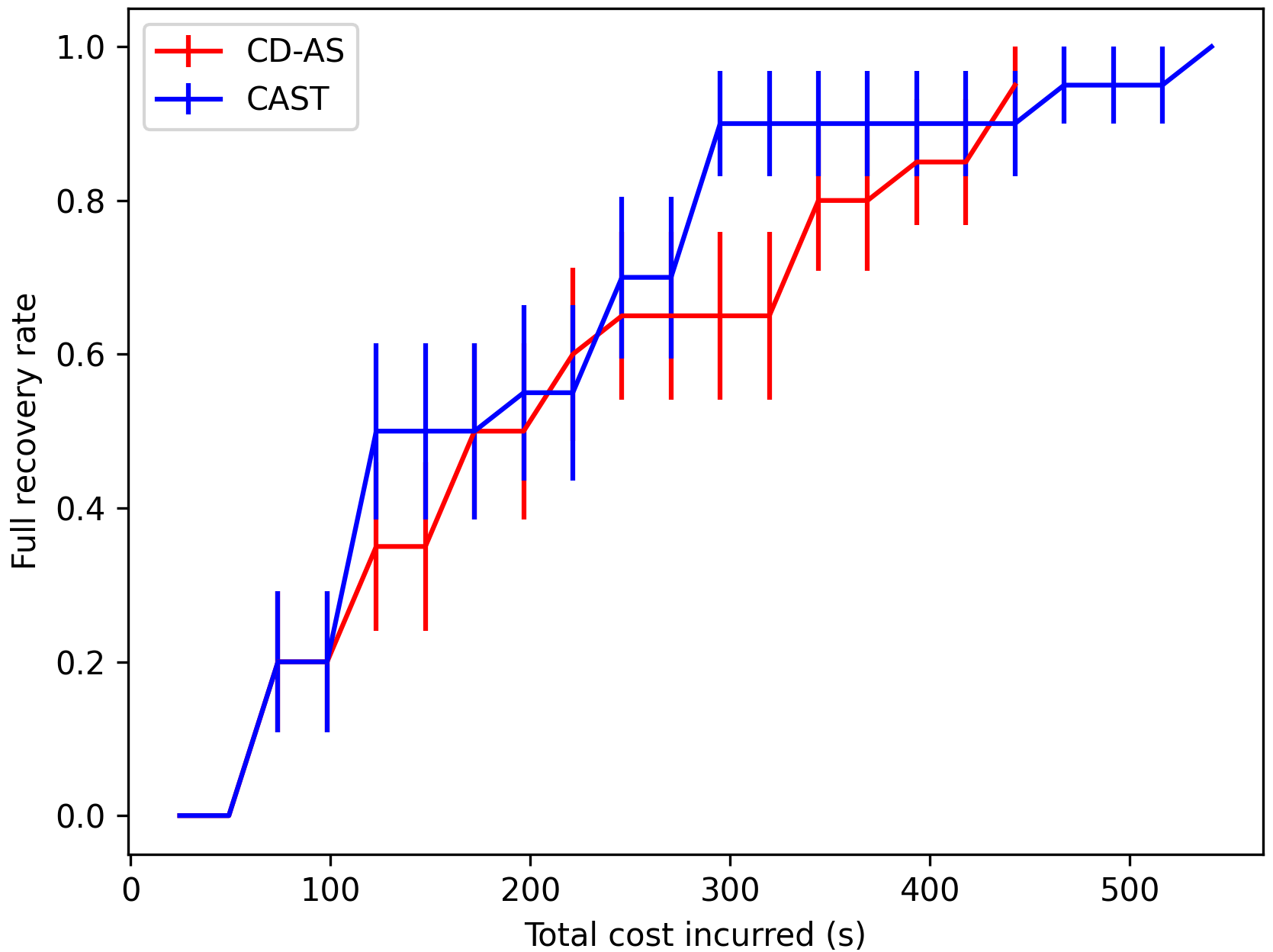}
    \caption{$\sigma=0.2\; c_s=50$s}
    \label{fig:1x16J1k1sigma0p2s50}
    \end{subfigure}
    \caption{\textbf{Cost-aware decision making in 1D search space of size $n=16$. $J=1$ agent. $k=1$ target. Low ($\sigma=\frac{1}{16}$) and high ($\sigma=0.2$) observation noise. }%
    Our diffusion based approach called CD-AS 
    incurs smaller or competitive 
    total cost compared to CAST when sensing cost is higher than traveling ($c_s=50$s). But when sensing cost is low, CAST selects sensing actions which incur a smaller total cost.  
    Plots show mean and standard error over 20 trials.} 
    \label{fig:genplanresults_1x16J1k1sigma1by16_costaware}
\end{figure}

In \cref{fig:genplanresults_1x16J1k1sigma1by16_costaware}, we compare the performance of online tree search (CAST) versus 
CD-AS with $T_{\text{diff}}=32$ for lookahead cost-aware decision making in active search under different cost scenarios. 
%
%
CAST solves a multi-objective optimization problem over travel and sensing cost, so when sensing cost is low ($c_s=0$s), 
agents minimize the total incurred travel cost and select successive sensing locations close to one another, possibly requiring more measurements than the optimal sensing sequence. 
In contrast, when sensing cost is much higher than travel cost (sensing cost $c_s=50$s), 
CAST agents optimize the number of measurements or sensing actions to minimize the incurred total cost. 
%
In \cref{fig:genplanresults_1x16J1k1sigma1by16_costaware}, when the observation noise is higher ($\sigma=0.2$), 
the total cost incurred for full recovery increases for both CAST and CD-AS since a particular grid cell might need to be sensed multiple times to avoid false positives and negatives. 
%
%
But when sensing cost is $c_s=0$s, CAST outperforms CD-AS even with $\sigma=0.2$. 
This is because the pareto-front constructed by CAST considers the ground truth total cost for a sequence of actions, whereas the cost-awareness in CD-AS is determined by the  
parameters $\alpha, \;\lambda$ and the 
learned distance function $d_\varphi$. 
We observed that the network $d_\varphi$ (when trained on the same dataset samples as $\nu_\psi$) has poor generalization over the combinatorial space of length $H = 8$ lookahead action sequences, which explains the weaker cost-awareness of CD-AS. 
On the other hand, when sensing cost is $c_s = 50$s and dominates the travel cost $c_d$ in the total estimated 
cost,  CD-AS outperforms CAST across both levels of observation noise, further supporting our observation in \cref{fig:genplanresults_1x16J1k1_myopicvslookahead} that diffusion based lookahead with gradient guidance recovers the optimal sequence of adaptive sensing actions.

\subsection{CD-AS in two-dimensional search space}
Next, we consider a two-dimensional (2D) search space discretized into $8\times8$ grid cells, with $J = 1$ agent and $k = 1$ targets. 
%
%
As before, we consider the same two levels of observation noise $\sigma$ and simulate an information-greedy myopic active search agent for gathering training data. 

We train the diffusion model on sequences of length $H=10$ 
%
with the previously described U-Net architecture for the trajectory network $\rho_\theta$ and $T_{\text{diff}} = 64$ diffusion denoising steps. 
%
For the return network $\nu_\psi$, we use the GNN architecture and optimize \cref{eq:diffusion_return_loss} with $T_{\text{diff}}=0$ i.e. without diffusion noising. 
We observe that this leads to better training sample efficiency and lower prediction error. 
For evaluation, we use a gradient guidance coefficient of $\alpha=10$. 
At each decision step, we sample $N_{\text{diff}} = 100$ action sequences of length $H=10$ conditioned on the current belief state, execute the first action and replan (re-sample) conditioned on the updated posterior belief at the next timestep (\cref{algo:genplan_gradguidance}). 
%




\begin{figure}[htp]
  \centering
    \begin{subfigure}{0.5\linewidth}
    \centering
    \includegraphics[width=\textwidth]{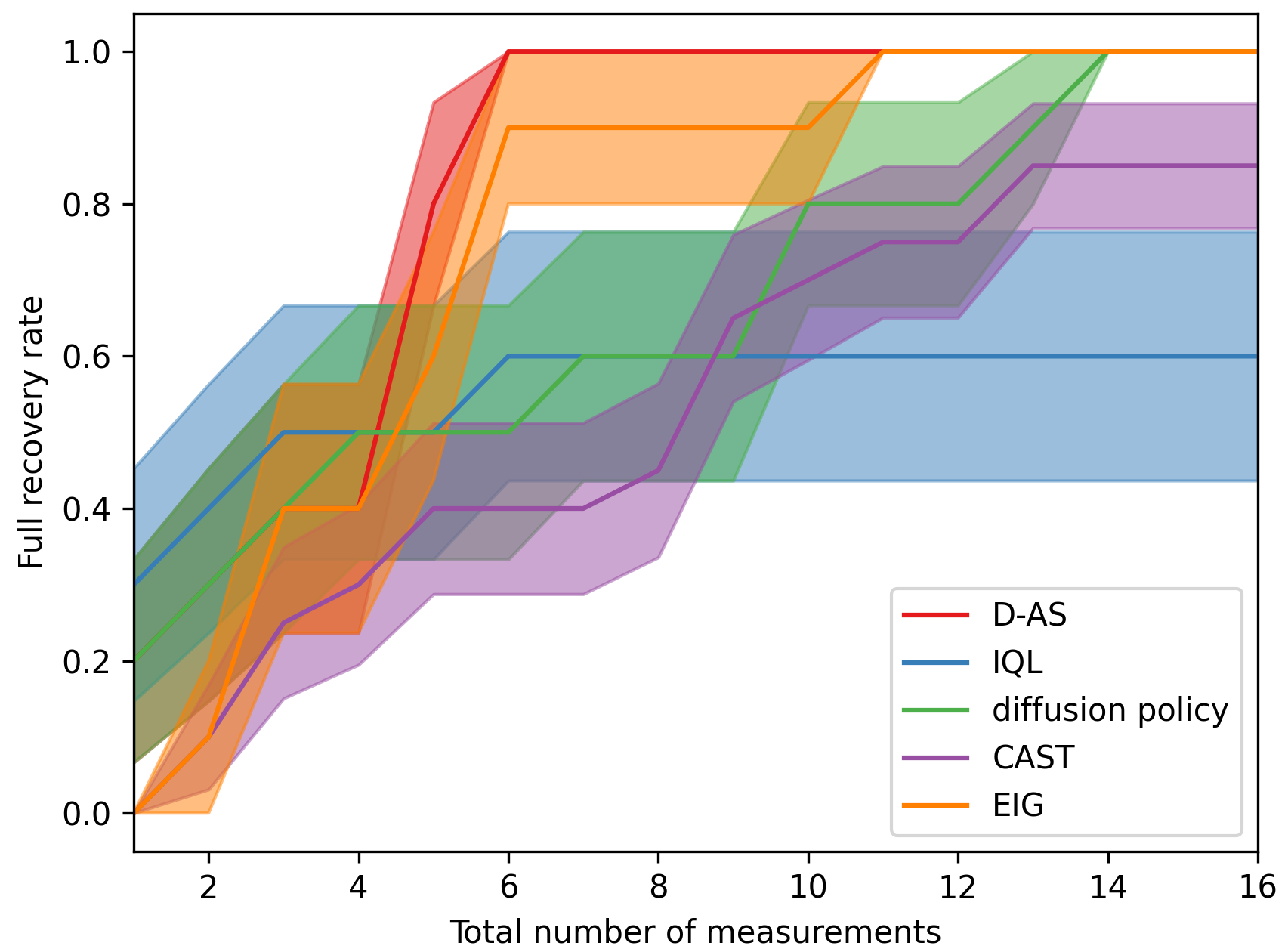}
    \caption{$\sigma=\frac{1}{16}$}
    \label{fig:8x8J1k1sigma1by16myopicfrec}
    \end{subfigure}%
    \begin{subfigure}{0.5\linewidth}
    \centering
    \includegraphics[width=\textwidth]{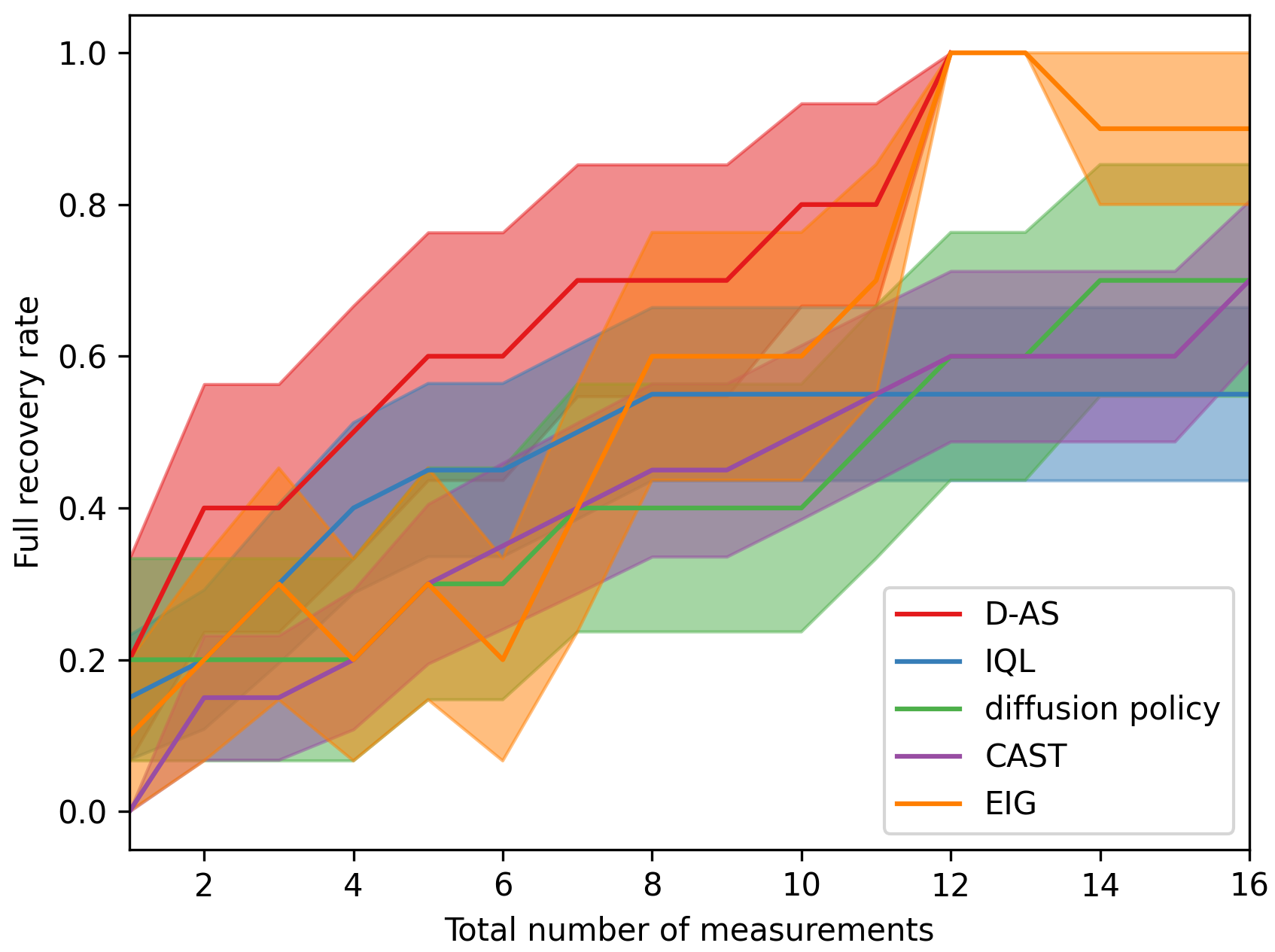}
    \caption{$\sigma=0.2$}
    \label{fig:8x8J1k1sigma0p2myyopicfrec}
    \end{subfigure}
    \caption{\textbf{Lookahead vs. myopic decision making in 2D search space of size $8\times8$. $J=1$ agent. $k=1$ target. Low ($\sigma=\frac{1}{16}$) and high ($\sigma=0.2$) observation noise. }%
    Our diffusion based approach (D-AS) samples the optimal sequence of actions and achieves full recovery with fewer measurements compared to myopic active search (EIG), offline RL (IQL), behavior cloning based diffusion (diffusion policy) and shallow online tree search (CAST) baselines. Plots show mean and standard error over 10 trials. } 
    \label{fig:genplanresults_8x8J1k1_myopicvslookahead}
\end{figure}

\cref{fig:genplanresults_8x8J1k1_myopicvslookahead} shows that our approach of diffusion based lookahead decision making (D-AS) outperforms myopic greedy as well as shallow online lookahead baselines. 
We do not focus on cost-awareness in this comparison ($\lambda=0$ in \cref{line:genplan_algogradguidance}). 
CAST is simulated with a budget of 25000 state-action rollouts to build a search tree of depth 2 at each online decision making step. 
We further discuss the trade-off between decision quality and time per decision making step for CAST in subsequent sections. 
%
%
%
Recall that EIG is the 
behavior policy used in simulations to collect the training data. 
Since EIG is a myopic greedy algorithm, it does not always recover the optimal action sequence. 
In spite of sub-optimal action sequences in the training data, our gradient-guided diffusion model 
is able to generate actions that recover the target with fewer measurements than the EIG policy. 
We also consider IQL as a baseline due to its reported competitive performance in offline RL benchmark environments that benefit from a multi-step dynamic programming type algorithm to stitch together sub-optimal sequences into optimal plans or trajectories \cite{kostrikovoffline}. 
Unfortunately, IQL does not consider stochastic policies or multimodality in action sequences for a task, which explains why IQL is unable to recover the target(s) in our active search setting. 
Diffusion policy is another baseline for decision making with diffusion but it follows the behavior cloning framework, hence its performance is limited by the quality of the training dataset. 
We observe that it is competitive with some of the active search baselines in the low noise setting but its performance deteriorates with higher observation noise. 
This is because the training dataset is collected with the EIG behavior policy, which is myopic in nature and does not support behavior cloning for lookahead decision making. 
In fact the lack of an optimal training dataset is a practical and important constraint for learning based approaches in long horizon planning or lookahead decision making with autonomous agents. 
The ability to utilize and learn from sub-optimal action sequences is an advantage of our gradient-guided diffusion framework as a generalizable and data-efficient active search algorithm. 

\begin{figure}[htp]
  \centering
    \begin{subfigure}{0.5\linewidth}
    \centering
    \includegraphics[width=\textwidth]{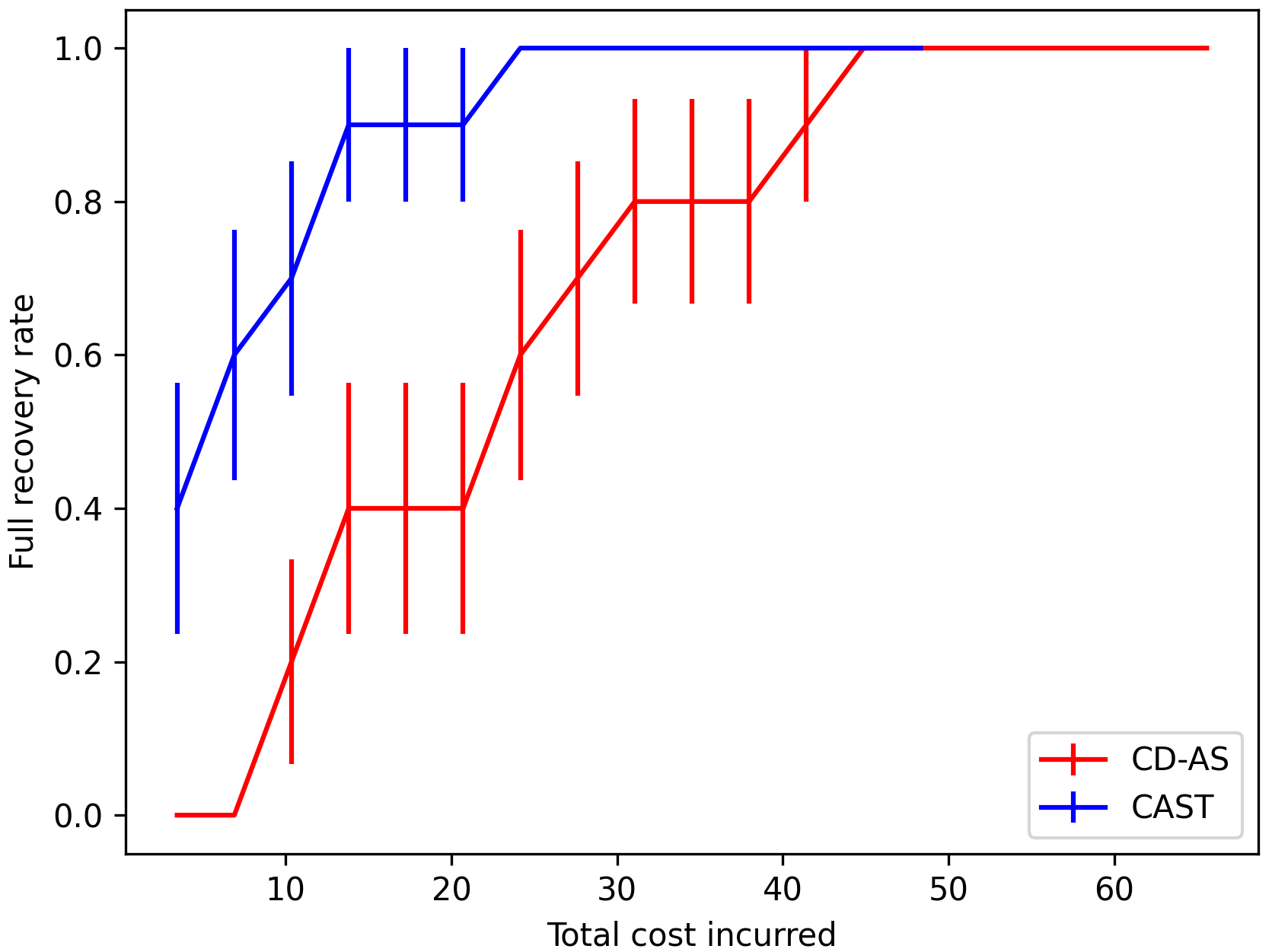}
    \caption{$c_s=0$s}
    \label{fig:8x8J1k1sigma1by16s0}
    \end{subfigure}%
    \begin{subfigure}{0.5\linewidth}
    \centering
    \includegraphics[width=\textwidth]{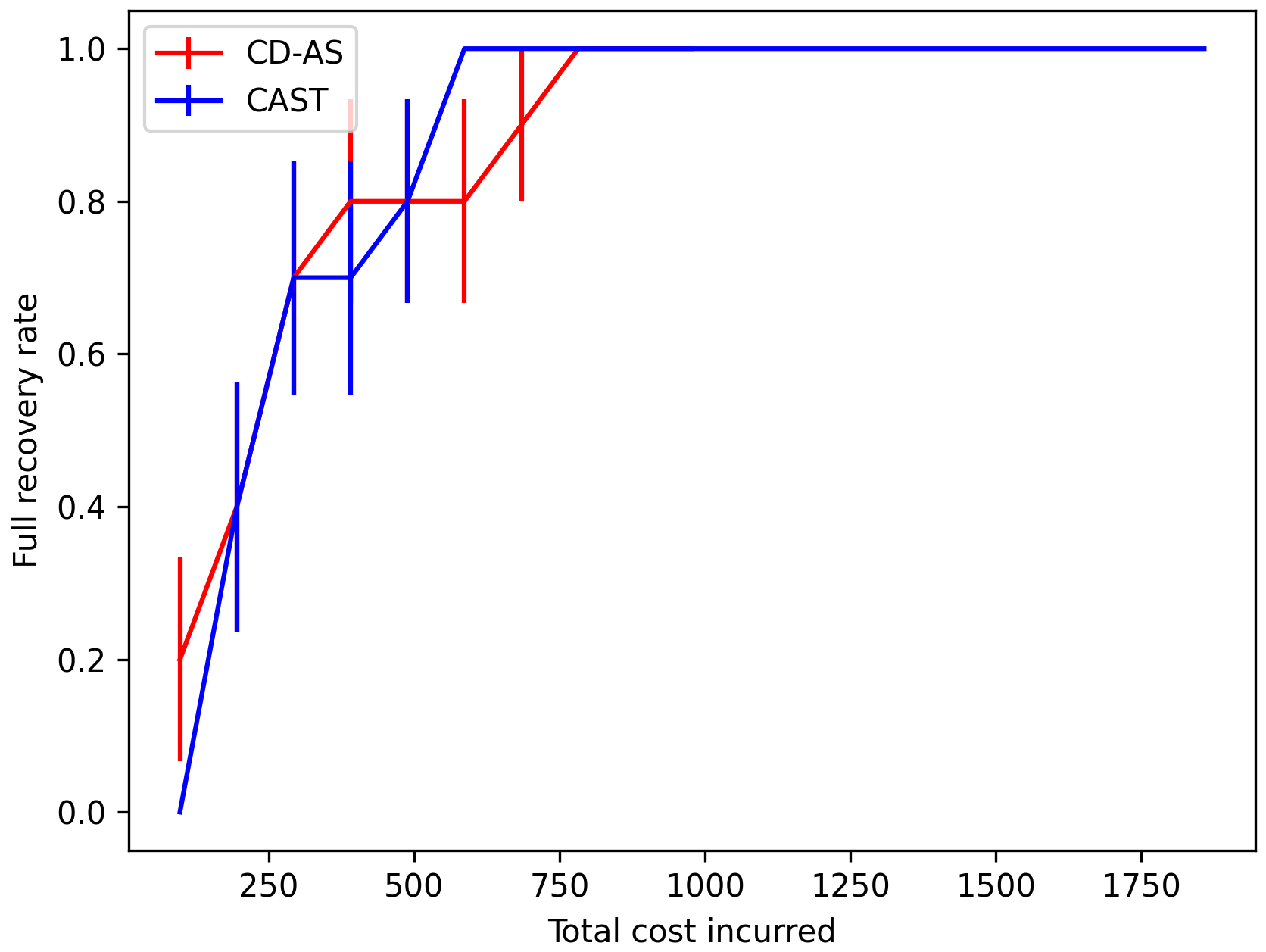}
    \caption{$c_s=50$s}
    \label{fig:8x8J1k1sigma1by16s50}
    \end{subfigure}
    \caption{\textbf{Cost-aware decision making in a 2D search space of size $8\times8$. $J=1$ agent. $k=1$ target. $\sigma=\frac{1}{16}$. }
    Our diffusion based approach CD-AS incurs a higher total cost in seconds compared to CAST which combines cost-optimization with tree search based planning. Plots show mean and standard error over 10 trials.} 
    \label{fig:genplanresults_8x8J1k1sigma1by16_costaware}
\end{figure}

In \cref{fig:genplanresults_8x8J1k1sigma1by16_costaware}, we compare the cost-aware performance of CD-AS ($T_{\text{diff}}=32$) with CAST. 
Recall that CAST computes the ground truth cost of sensing actions during tree search based online decision making, whereas 
cost-awareness in CD-AS is incorporated by gradient guidance from the trained distance network $d_\varphi$ during denoising diffusion sampling. 
We observe that 
%
CAST outperforms CD-AS under both cost scenarios. 
Spatial distance estimation with neural networks is a difficult learning problem \cite{ramakrishnan2024does}, therefore we observe that gradient guidance with a scalarized combination of $\nu_\psi$ and $d_\varphi$ does not sample cost-aware action sequences as effectively as constructing a pareto-front with ground truth cost estimates. 

\subsection{CD-AS in multi-agent active search}

\begin{figure}[htp]
  \centering
  \includegraphics[width=0.25\textwidth]{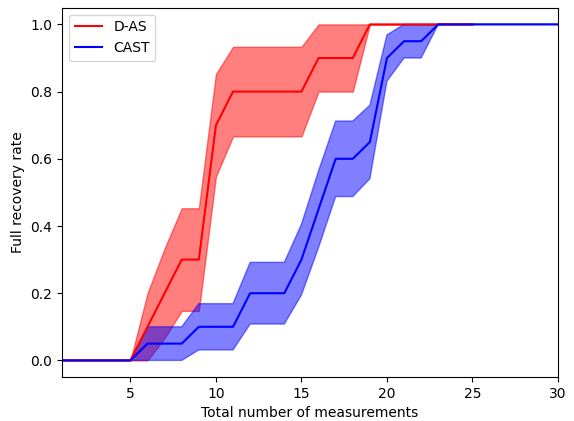}
  \caption{\textbf{Lookahead vs myopic decision making in 2D $8\times8$ search space. $J=3$ agents. $k=4$ targets. $\sigma=\frac{1}{16}$. }
    Our diffusion based approach recovers the optimal sequence of actions for full recovery faster compared to shallow lookahead multi-agent active search with CAST. Plots show mean and standard error over 10 trials.}
  \label{fig:genplan_j3k4_fullrecvsT}
    %
\end{figure}

\begin{figure}[htp]
  \centering
    \begin{subfigure}{0.5\linewidth}
    \centering
    \includegraphics[width=\textwidth]{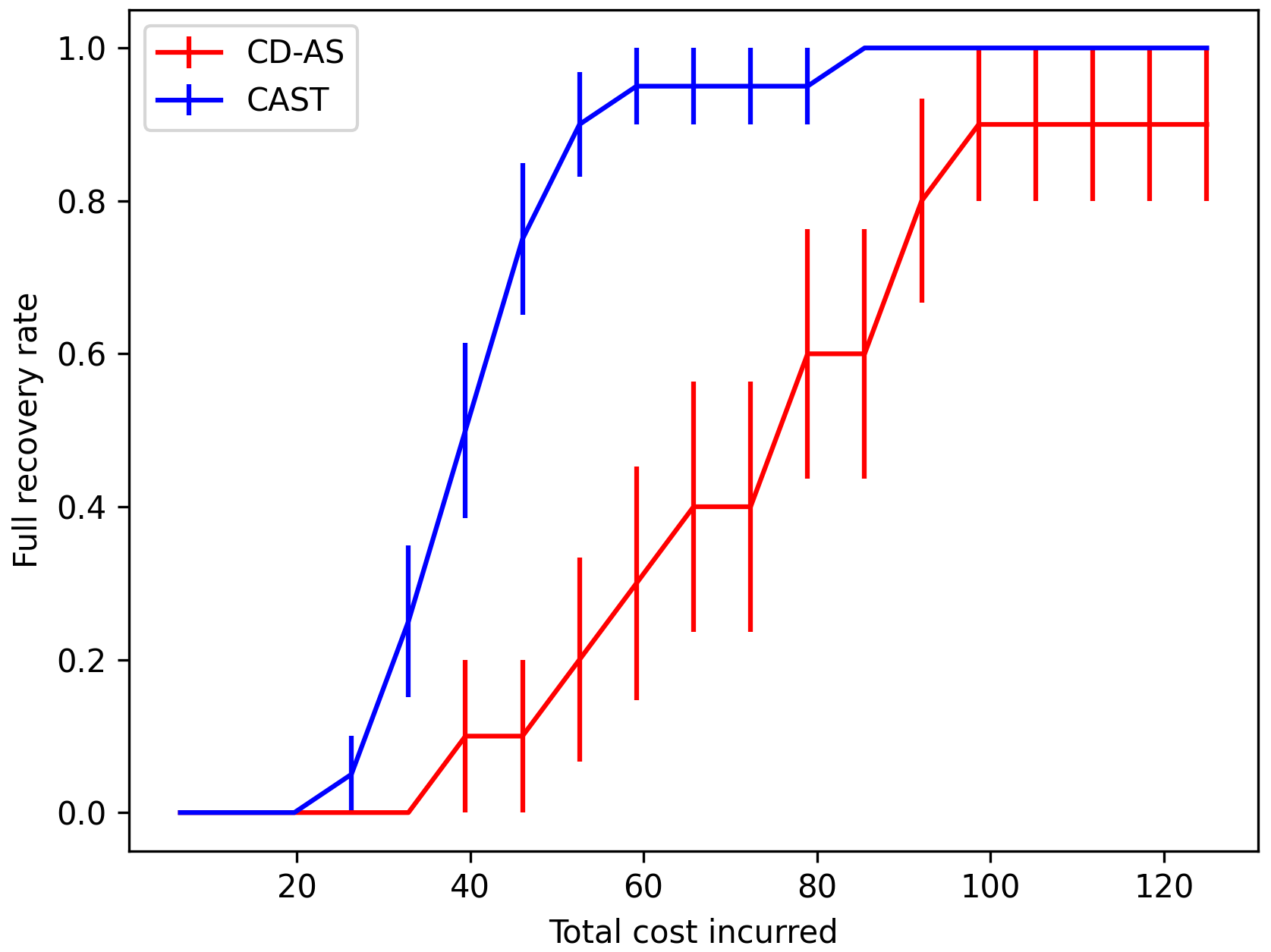}
    \caption{$c_s=0$s}
    \label{fig:8x8J3k4sigma1by16s0}
    \end{subfigure}%
    \begin{subfigure}{0.5\linewidth}
    \centering
    \includegraphics[width=\textwidth]{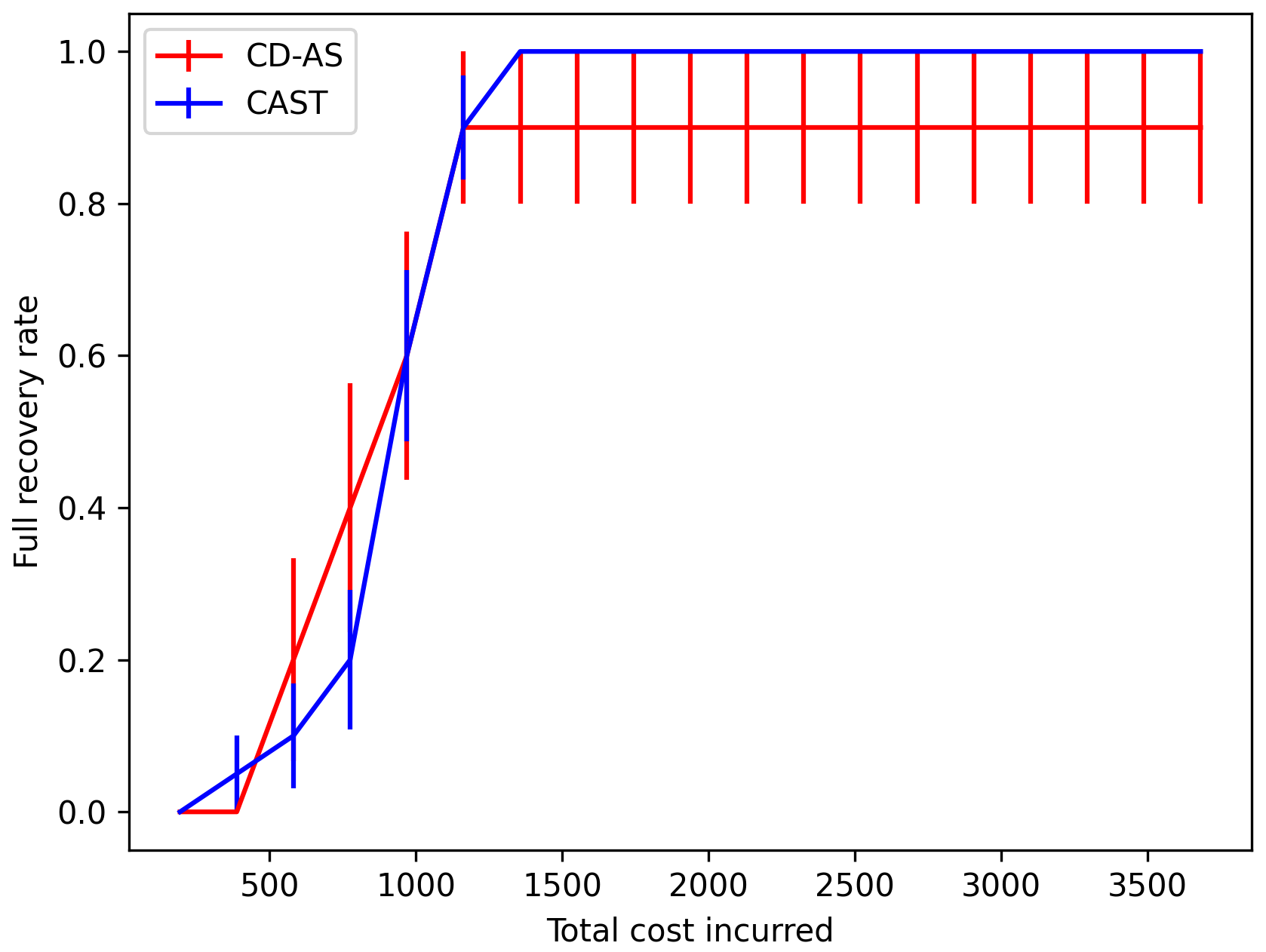}
    \caption{$c_s=50$s}
    \label{fig:8x8J3k4sigma1by16s50}
    \end{subfigure}
    \caption{\textbf{Cost-aware decision making in 2D search space of size $8\times8$ . $J=3$ agents. $k=4$ targets. $\sigma=\frac{1}{16}$. }%
    Our diffusion based approach CD-AS 
    incurs a higher 
    total cost compared to CAST which combines ground truth cost-awareness with search tree based planning. Plots show mean and standard error over 10 trials.} 
    \label{fig:genplanresults_8x8J3k4_costaware}
\end{figure}

We consider $J=3$ agents and $k=4$ targets distributed in an $8\times8$ search space. 
%
Following \cref{algo:genplan_gradguidance_multiagent}, each agent samples its own set of action sequences conditioned on its belief state. 
In contrast, multi-agent CAST relies on Thompson sampling to build each agent's search tree. 
We observe that in \cref{fig:genplan_j3k4_fullrecvsT}, our diffusion based approach D-AS (with $\lambda=0$) 
outperforms CAST in terms of the number of measurements required to completely recover all targets. 
\cref{fig:genplanresults_8x8J3k4_costaware} further compares their cost-awareness. 
When sensing is more expensive than traveling (sensing cost $c_s=50$s), CD-AS is competitive with CAST in terms of the total cost incurred for full recovery. 
But CAST outperforms CD-AS when sensing cost is low ($c_s=0$s). 
In the latter case, CAST optimizes for the ground truth travel cost incurred in active search and selects action sequences that incur a low travel cost but do not optimally explore the search space. In contrast, CD-AS does not sample such low cost sequences since the training dataset was collected with the EIG policy which is not cost-aware and does not trade-off cost of sensing actions with information gain from exploring the search space. 
%
%

\subsection{Comparing the computational cost of generative planning 
with lookahead decision making}
\label{subsec:genplan_timecompare}


Our proposed approach CD-AS for cost-aware gradient-guided diffusion is also motivated by the goal of improving the time complexity of decision making 
in lookahead active search. 
In \cref{tab:genplan_timeperdecision} we compare the CPU wall clock time per decision making step in CAST with the GPU time for sampling lookahead action sequences and selecting the next cost-aware sensing action in CD-AS. 
For a 1D search space of size $n=16$, at every decision step, CAST constructs a search tree of depth 2 with 5000 rollouts for a measured wall clock time of 120.40 secs. 
In contrast, CD-AS takes 78.68 secs to sample a batch of 10000 action sequences of lookahead depth $H=8$ using $T_{\text{diff}}=32$ denoising steps. 
In the 2D search space of size $8\times 8$, while CAST requires 405.47 secs to build a search tree of depth 2 with 25000 rollouts, it only takes 132.75 secs for CD-AS to sample a batch of 100 action sequences with lookahead depth $H=10$ using $T_{\text{diff}}=32$ denoising steps. 
In both settings, we observe that denoising diffusion sampling in CD-AS allows for longer  $H$ rollouts at a lower time complexity. 
We believe that with 
future work in improving diffusion training and sampling techniques, the inference time 
for CD-AS can be further 
reduced thereby providing an efficient alternative approach to tree search for planning and lookahead decision making in active search.

\begin{table}[htp]
    \centering
    \begin{tabular}{ccc}
     $n_\ell\times n_w$    & CAST & CD-AS \\
     $1\times16$ &  120.40 & 78.68\\
     $8\times8$ & 405.47 & 132.75\\
    \end{tabular}
    \caption{\textbf{Comparing the average wall-clock time in seconds per decision making step with CAST and CD-AS in both 1D and 2D search spaces.} CD-AS amortizes the time complexity of rolling out state-action sequences in tree search 
    and leads to more than $30\%$ improvement per lookahead decision step.}
    \label{tab:genplan_timeperdecision}
\end{table}

\bibliographystyle{IEEEtran}
\bibliography{IEEEabrv,ref}

\end{document}